\documentclass[lettersize,journal]{IEEEtran}

\usepackage{tikz,xcolor,hyperref}
\definecolor{lime}{HTML}{A6CE39}
\DeclareRobustCommand{\orcidicon}{%
    \begin{tikzpicture}
    \draw[lime, fill=lime] (0,0) 
    circle [radius=0.16] 
    node[white] {{\fontfamily{qag}\selectfont \tiny ID}};    \draw[white, fill=white] (-0.0625,0.095) 
    circle [radius=0.007];    \end{tikzpicture}
    \hspace{-2mm}}
\foreach \x in {A, ..., Z}{%
    \expandafter\xdef\csname orcid\x\endcsname{\noexpand\href{https://orcid.org/\csname orcidauthor\x\endcsname}{\noexpand\orcidicon}}
    }

\usepackage{amsmath,amsfonts}
\usepackage[lined,boxed,ruled, commentsnumbered]{algorithm2e}
\usepackage{array}
\usepackage[caption=false,font=normalsize,labelfont=sf,textfont=sf]{subfig}
\usepackage{textcomp}
\usepackage{stfloats}
\usepackage{url}
\usepackage{verbatim}
\usepackage{graphicx}
\usepackage{cite}
\usepackage{amsmath}
\usepackage{amssymb}
\usepackage{mathtools}
\usepackage{amsthm}
\usepackage{hyperref}
\usepackage{url}
\usepackage[utf8]{inputenc} 
\usepackage[T1]{fontenc}    
\usepackage{hyperref}       
\usepackage{url}            
\usepackage{booktabs}       
\usepackage{amsfonts}       
\usepackage{nicefrac}       
\usepackage{microtype}      
\usepackage{xcolor}         
\usepackage{graphicx}
\usepackage{wrapfig}
\usepackage{multirow}
\usepackage{soul}
\usepackage{subfig}
\usepackage{comment}
\usepackage{threeparttable}
\usepackage{float}
\usepackage{bm}

\begin{document}

\title{DREAM: Domain-free Reverse Engineering Attributes of Black-box Model}

\author{Rongqing~Li\orcidA{}, Jiaqi~Yu\orcidB{}, Changsheng~Li\orcidC{},~\IEEEmembership{Member,~IEEE,} Wenhan~Luo\orcidD{},~\IEEEmembership{Senior Member,~IEEE,}\\ Ye~Yuan\orcidE{},~\IEEEmembership{Member,~IEEE,}~and~Guoren~Wang\orcidF{}

\thanks{Rongqing Li, Jiaqi Yu, Changsheng Li, Ye Yuan, and Guoren Wang are with the School of Computer Science and Technology, Beijing Institute of Technology, Beijing 100081, China (e-mail: lirongqing99@gmail.com; jiaqiyu@bit.edu.cn; lcs@bit.edu.cn; yuan-ye@bit.edu.cn; wanggrbit@126.com).}

\thanks{Wenhan Luo is with the School of Cyber Science and Technology, Shenzhen Campus of Sun Yat-sen University, Shenzhen, Guangdong 518107, China (e-mail: whluo.china@gmail.com).}

\thanks{Rongqing Li and Jiaqi Yu contribute equally to this work.}
\thanks{Changsheng Li is the corresponding author.}
}


\IEEEpubid{0000--0000/00\$00.00~\copyright~2021 IEEE}

\maketitle

\begin{abstract}
Deep learning models are usually black boxes when deployed on machine learning platforms. Prior works have shown that the attributes ($e.g.$, the number of convolutional layers) of a target black-box neural network can be exposed through a sequence of queries. There is a crucial limitation: these works assume the dataset used for training the target model to be known beforehand and leverage this dataset for model attribute attack. However, it is difficult to access the training dataset of the target black-box model in reality. Therefore, whether the attributes of a target black-box model could be still revealed in this case is doubtful. In this paper, we investigate a new problem of Domain-agnostic Reverse Engineering the Attributes of a black-box target Model, called DREAM, without requiring the availability of the target model's training dataset, and put forward a general and principled framework by casting this problem as an out of distribution (OOD) generalization problem. In this way, we can learn a domain-agnostic model to inversely infer the attributes of a target black-box model with unknown training data. This makes our method one of the kinds that can gracefully apply to an arbitrary domain for model attribute reverse engineering with strong generalization ability. Extensive experimental studies are conducted and the results validate the superiority of our proposed method over the baselines.
\end{abstract}


\section{Introduction}
\label{sec:intro}
\begin{figure*}
\setlength{\belowcaptionskip}{-0.4cm} 
  \begin{center}
  \includegraphics[width=1\linewidth]{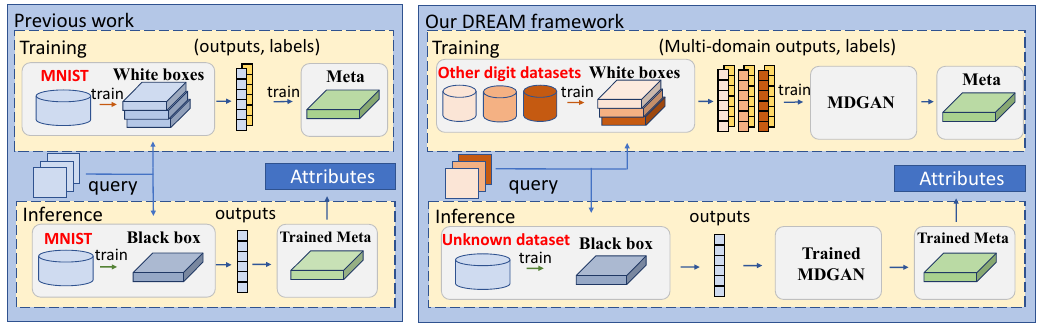}
  \caption{
      Previous work (left) assumes the dataset used to train the target the black-box model is given beforehand, and requires to use the same dataset to train white-box models.
      Our DREAM framework (right) relaxes the condition that training data of a black-box model is no longer required to be available. Our idea is to cast the task of the black-box model attribute inference into an out-of-distribution learning problem.
     } 
  \label{fig:intro_fig}
  \end{center}
\end{figure*}

\IEEEPARstart{W}{ith} its commercialization, machine learning as a service (MLaaS) is becoming more and more popular, and providers are paying more attention to the privacy of models and the protection of intellectual property.
Generally speaking, the machine learning service deployed on the cloud platform is a black box, where users can only obtain outputs by providing inputs to the model. 
The model's attributes such as architecture, training set, and training method, are concealed by the provider.
However, whether the deployment is safe? Once the attributes of the model are revealed, it will be beneficial to many downstream attacking tasks, e.g., adversarial example generation \cite{moosavi2016deepfool,li2022decision,doan2022tnt,aafaq2022language}, model inversion \cite{he2019model,khosravy2022model,zhu2022label}, etc. The work in \cite{oh2018towards2} has conducted model reverse engineering to reveal model attributes, as shown in the left of Fig. \ref{fig:intro_fig}. They first collect a large set of white-box models which are trained based on the same datasets as the target black-box model, $e.g.$, the MNIST hand-written dataset \cite{lecun_mnist_1998}.
Given a sequence of input queries, the outputs of white-box models can be obtained. After that, a meta-classifier is trained to learn a mapping between model outputs and model attributes. For inference, outputs of the target black-box model are fed into the meta-classifier to predict model attributes. The promising results demonstrate the feasibility of model reverse engineering.
However, a crucial limitation in \cite{oh2018towards2} is that they assume the dataset used for training the target model to be known in advance, and leverage this dataset for meta-classifier learning.
In most application cases, the training data of a target black-box model is unknown. 
When the domain of training data of the target black-box model is inconsistent with that of the set of constructed white-box models, the meta-classifier is usually unable to generalize well on the target black-box model. 
\begin{figure}
\setlength{\belowcaptionskip}{-0.4cm} 
  \begin{center}
  \includegraphics[width=0.9\linewidth]{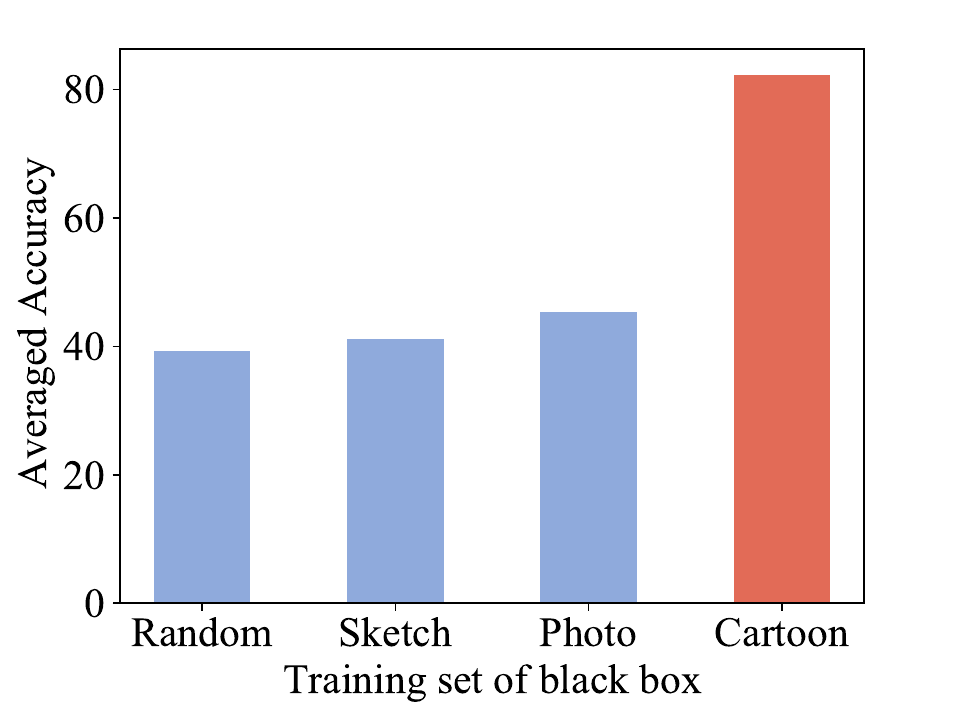}
  \caption{
    The performance of KENNEN\cite{oh2018towards2}
 on black-box model trained on Cartoon, Sketch and Photo dataset \cite{li2017deeper}. The meta-classifier in KENNEN is trained on white-box models that are trained on the Cartoon dataset.
}
  \label{fig:intro_2}
  \end{center}
  \vspace{-2em}
\end{figure}
To verify this point, we train three black-box models with the same architecture on three different datasets, Photo, Cartoon and Sketch\cite{li2017deeper}, respectively.
We use the method in \cite{oh2018towards2} to train a meta-classifier on the white-box models which are trained on the Cartoon dataset. After that, we use the trained meta-classifier to infer attributes of three black-box models, respectively.
\IEEEpubidadjcol
As shown in Fig. \ref{fig:intro_2}, when the training dataset of black-box models and white-box models are the same (i.e., Cartoon), the performance reaches about $80\%$. Otherwise, it sharply drops to about $40\%$, close to random guess. The huge gap shows that it is not trivial to investigate model reverse engineering with the assumption of the training dataset of the black-box model is not available. Furthermore, if the training set for the black-box model changes, \cite{oh2018towards2} needs to retrain the whole set of white-box models to obtain a promising result, which is extremely time-consuming.

In this paper, we investigate the problem of black-box model attribute reverse engineering, no longer requiring the availability of training data of the target model, as shown in the right of Fig. \ref{fig:intro_fig}.
Obviously, when feeding the same input queries to models with the same architecture but training data of different domains, the output distributions of these models are usually different. 
Thus, in our problem setting, a key point is how to bridge the gap between the output distributions of white-box and target black-box models, due to the lack of the target model's training data. An ideal meta-classifier should be well trained based on the outputs of white-box models, and predict well on outputs of the target black-box model, even if white-box and black-box models are trained using data of different domains.

In light of this, we cast such a problem as an out-of-distribution (OOD) generalization problem, and propose a novel framework DREAM: Domain-agnostic Reverse Engineering the Attributes of black-box Model.
In the field of computer vision, OOD generalization learning has been widely studied in recent years \cite{shen2021towards,yi2021improved,ye2021towards,zhou2022domain}, where its main goal is to learn a model on data from one or multiple domains and to generalize well on data from another domain that has not been seen during training. 
One kind of  mainstream OOD learning approach is to extract domain invariant features from data of multiple different domains, and utilize the domain invariant features for downstream tasks \cite{li_domain_2018,kim2021selfreg,zhou2021mixstyle,li2021learning,matsuura2020domain,jia2020single}. 
However, these methods mainly focus on image or video data and have shown powerful performance. 
Back to our problem, the black-box models deployed on the cloud platform usually provide which categories they can output. Therefore,  we can collect data with the same label but different distributions as domains to train white-box models and obtain their probability outputs. \textcolor{black}{These outputs of different domains will be utilized for OOD learning.}
Since the data we concentrate on is related to the outputs of machine learning models, $e.g.$, probability values, how to design an effective OOD learning method over this type of data has not been explored. 
To this end, we introduce a multi-discriminator generative adversarial network to learn domain-invariant features from the outputs of white-box models trained on multi-domain data.
Based on learned domain-invariant features, we learn a domain-agnostic reverse model, which can well infer the attributes of a target black-box model trained on data of an arbitrary domain.
 
Our contributions are summarized as follows:
1) We provide the first study on the problem of domain-agnostic reverse engineering the attributes of black-box models and cast it as an OOD generalization problem;
2) We propose a generalized framework, DREAM, which can address the problem of inferring the attributes of a black-box model with an arbitrary training domain;
3) We constitute the first attempt to explore learning domain invariant features from probability representations, in contrast to traditional image representations;
4) We perform extensive experiments, demonstrating the effectiveness of our method.

\section{Related Works}
\label{related_works}
\textbf{Reverse Engineering of Model Attribute.} Its goal is to reveal attribute values of a target model, such as model structure, optimization method, hyperparameters, $etc$.Current research efforts focus on two aspects, hardware \cite{yan_cache_2020,hua_reverse_2018,zhu_hermes_2021} and software \cite{oh2018towards2,wang_stealing_2019,rolnick2020reverse,asnani2021reverse}. The hardware-based methods utilize information leaks from side-channel \cite{hua_reverse_2018,yan_cache_2020} or unencrypted PCIe buses \cite{zhu_hermes_2021} to invert the structure of deep neural networks. Software-based methods reveal model attributes by machine learning. \cite{wang_stealing_2019} steals the trade-off weight of the loss function and the regularization term. They derive over-determined linear equations and solve the hyperparameters by the least-square method. 
\textcolor{black}{\cite{asnani2021reverse} theoretically proves the weight and bias can be reversed in linear network with ReLU activation. \cite{rahman_correlation-aware_2020} infers hyperparamters and loss functions of generative models through the generated images.}
KENNEN \cite{oh2018towards2} prepares a set of white-box models and then trains a meta-classifier to build a mapping between model outputs and their attributes. It is the most related work to ours. However, a significant difference is that KENNEN \cite{oh2018towards2} requires the data used to train the target black-box model to be given beforehand. Our method relaxes this condition, $i.e.$, we no longer require the training data of the target model to be available, which is a more practical problem.

\textbf{Model Functionality Extraction.} It aims to train a clone model that has similar model functionality to that of the target model. To achieve this goal, many works have been proposed in recent years  \cite{orekondy_knockoff_2019,truong_data-free_2021,papernot_practical_2017,kariyappa_maze_2021,wang2022dst,sanyal2022towards,wang2021delving}. \cite{orekondy_knockoff_2019} uses an alternative dataset collected from the Internet to query the target model. \cite{papernot_practical_2017} assumes part of the dataset is known and then presents a dataset augmentation method to construct the dataset for querying the target model. \textcolor{black}{Moreover, data-free extraction methods \cite{kariyappa_maze_2021,truong_data-free_2021,papernot_practical_2017,wang2022dst,sanyal2022towards,wang2021delving} query a target model through data generated by a generator, without any knowledge about the training data distribution. Different from the methods mentioned above, our goal is to infer the attributes of a black-box model, rather than stealing the model function.}

\textbf{Membership Inference.} Its goal is to determine whether a sample belongs to the training set of a model \cite{he2020segmentations,choquette2021label,rezaei2021difficulty,9833649,10.1145/3548606.3560675,10.1145/3548606.3560684}. Although   inferring model attribute is different from the task of membership inference, the technique in \cite{oh2018towards2} is actually similar to those of membership inference attack. However, as stated aforementioned, when the domain of training data of the target black-box model is inconsistent with that of the set of white-box models, the method is usually unable to generalize well because of the OOD problem.

\textbf{OOD Generalization.} The goal of OOD Generalization is to deal with the inevitable shifts from a training distribution to an unknown testing distribution \cite{shen2021towards}. Existing methods mainly fall into three categories: domain generalization \cite{kim2021selfreg,li_domain_2018,zhou2021mixstyle,zhou_domain_2021,hu_domain_2020}, causal learning \cite{arjovsky2019invariant,creager2021environment,krueger2021out,mahajan2021domain} and stable learning \cite{shen2020stable,kuang2020stable,zhang2021deep,kuang2018stable}. Domain generalization attempts to learn invariant representations among different domains. Causal learning and stable learning aim to search for causal features to ground-truth labels from data and filter out label-unrelated features. The former makes existing causal features invariant, while the latter focuses on the effective features strongly related to labels by reweighting attention. In addition, the DANN \cite{ganin2016domain} adopts generative adversarial network to solve domain adaptation task. Since the above methods mainly focus on images or videos, how to design an effective OOD learning method for attribute inference of the black-box model has not been explored so far.

\section{Proposed Methods}
 \begin{figure*}
  \begin{center}
  \includegraphics[width=1.0\linewidth]{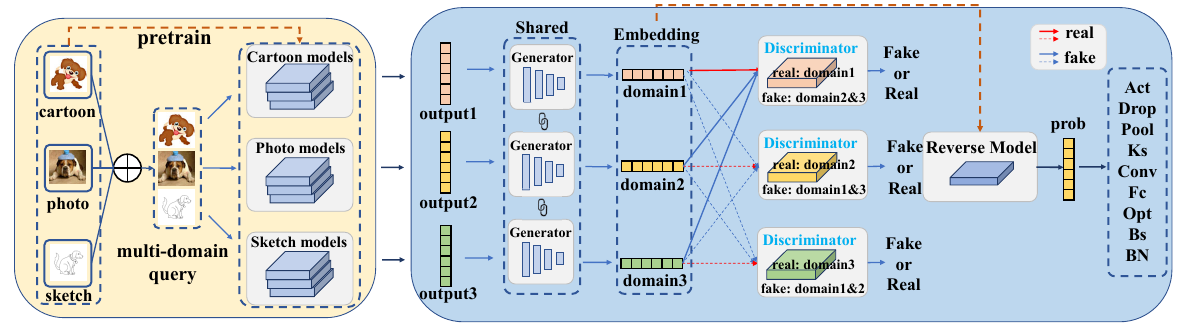}
  \caption{ An illustration of our DREAM framework.
      1) In the left part, we prepare a large number of white-box models from different domains. In each domain, models have various combinations of attributes, then we input multi-domain queries into each white-box model to obtain a multi-domain model output. 2) In the right part, we introduce a multiple-discriminator GAN network to learn domain invariant features from the probability outputs of white-box models trained on multi-domain data. After that, a domain-agnostic reverse model is learned based on domain invariant embeddings and is used to infer the attributes of the black-box model with arbitrary domains.
  }
  \label{overall}
  \end{center}
\end{figure*}
\subsection{Preliminaries}
\label{sec:preliminaries}
\textbf{KENNEN \cite{oh2018towards2}.} Given a black-box model $B$, model attribute reverse engineering in \cite{oh2018towards2} aims to build a meta-classifier $\Phi: B \rightarrow A$, where $A$ is the set of model attributes including model architecture, optimizer, and training hyperparameters, $etc$.
Concretely, they first collect a large set of white-box models $\mathcal{F}$ containing different attributes combination and train these white-box models based on the same training data $\mathcal{D}$ as  that of the target black-box model. Then outputs $O$ are obtained by querying these white-box models with a sequence of input images. Finally, they build a mapping $\Phi$ from outputs $O$ to model attributes $A$, realized as a meta-classifier $\Phi$.
At the inference phase, the meta-classifier takes outputs from the target model as input and predicts the corresponding attributes.
 
\textbf{Problem Formulation.} As aforementioned, there is a strict constraint in  \cite{oh2018towards2} that they assume the training dataset $\mathcal{D}$ of the target model to be given in advance, and leverage $\mathcal{D}$ for learning meta-classifier $\Phi$. 
In most scenarios, especially on public machine learning platforms, it is difficult to access the  training data of a target black-box model, which significantly limits the applications of \cite{oh2018towards2}.
To mitigate this problem, we provide a new problem setting by relaxing the above constraint, $i.e.$, we no longer require the training data $\mathcal{D}$ of the target black-box model to be available. Thus, our goal is to learn a domain-agnostic reverse classifier $\Phi$ that  is trained based on outputs of white-box models $\mathcal{F}$, and predict well for the target black-box model, even if white-box and black-box models are built based on training data of different domains. 

\subsection{DREAM Framework}
To perform domain-agnostic black-box model attribute reverse engineering, we cast this problem into an out-of-distribution (OOD) generalization learning problem, and propose a novel framework DREAM, as shown in Fig. \ref{overall}.
Our DREAM framework consists of two parts:
In the left part of Fig. \ref{overall}, we train a number of white-box models with training sets from different domains. 
Models of each domain are enumerated with various model attributes.
All of these models constitute a model set covering different domains (please refer to Sect. \ref{subsec:dataset_construction} for more details). 
Next, we prepare queries as input to these models.
For each domain, we sample an equal number of images from the corresponding dataset and concatenate them as a batch of queries. 
These queries are sent to each model, and outputs of the model are fed into the other module of our DREAM framework,  as shown in the right part of Fig. \ref{overall}. To learn domain invariant features, we introduce a multi-discriminator generative adversarial network, where consists of multiple discriminators corresponding to different domains and one generator across multiple domains. The generator aims to learn domain invariant features, and  each discriminator intends to make the learned feature distributions of other domains fit that of the domain itself.  
In this way, the generator is capable of learning domain invariant features.
Based on the learned domain invariant features, we further learn a domain-agnostic reverse model to infer the attributes of a black-box model with an arbitrary domain.

\subsection{Multi-domain Outputs Preparation}
The multi-domain output can be taken as a representation of a white-box model, and is fed into the multi-discriminator generative adversarial network to learn domain invariant features.
Specifically, we sample an equal number of images from the dataset of each domain to obtain a query set  $Q = \{q_j\}_{j=1}^N$, where $N$ is the number of total queries. 
We denote training model set from each domain as $\mathcal{F} = [\mathbf{f^1}, \mathbf{f^2}, ..., \mathbf{f^m}]$, where $\mathbf{f^i}$ is model of $i^{th}$ domain.
Then we obtain output $\{O_j^i\}_{j=1}^N$ for each domain by querying $\mathbf{f^i}$ and concatenate outputs of the same domain ($N$ instances in total) together to make 1-d vector $O^i \in \mathbb{R}^{CN}$. Each 1-d vector is a representation of model, and $C$ is the number of classes in the dataset. 
Finally, we derive multi-domain outputs as $O = [O^1,...,O^m] \in \mathbb{R}^{m \times CN}$.

The multi-discriminator generative adversarial network aims to learn embeddings for probability outputs of each domain by a parameter-sharing generator, and make the distributions of different domains as close as possible by multiple discriminators.

\subsection{Multi-Discriminator GAN}
After preparing multi-domain probability outputs, we introduce a GAN-based network \cite{goodfellow2020generative}, with the purpose of learning domain invariant features from the probability outputs of white-box models trained on multi-domain representation.

 \begin{figure}
  \begin{center}
  \includegraphics[width=1\linewidth]{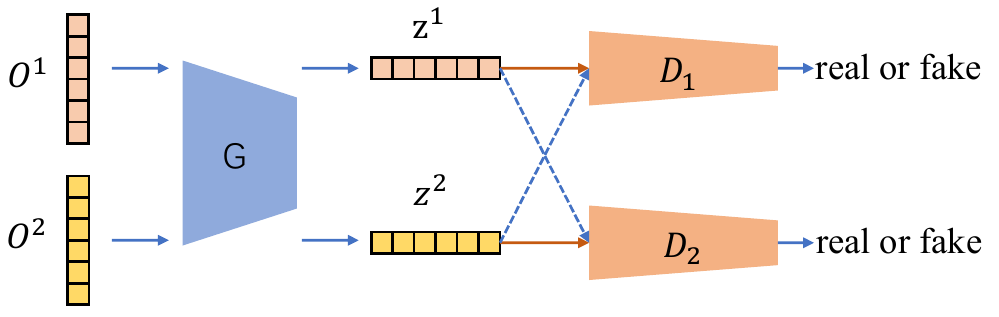}
  \caption{An example to illustrate the multi-discriminator GAN.}
  \label{mdgan}
  \end{center}
\end{figure}

To better present, we take Fig. \ref{mdgan} to illustrate how the multi-discriminator GAN works. Assume there are two kinds of inputs, $O^1$ and $O^2$, from two domains. When feeding them into the generator $G$, we can obtain the corresponding embeddings $z^1$ and $z^2$, respectively. After that, we feed $z^1$ and $z^2$ to the discriminator $D_1$, where $D_1$ is expected to output a ``real" label for $z^1$ and output a ``fake" label for $z^2$. By jointly training $G$ and $D_1$ based on a min-max optimization, the distribution of $z^2$ is expected to move towards that of $z^1$.  In the meantime, we also feed $z^1$ and $z^2$ to the discriminator $D_2$. Differently, $D_2$ is expected to output a ``real" label for $z^2$ and output a ``fake" label for $z^1$. By jointly training $G$ and $D_2$, the distribution of $z^1$ is expected to move towards that of $z^2$.
In this way,  $z^1$ and $z^2$ generated by the generator $G$ become domain invariant representations.

Formally, we define $G(O; \theta_g): O \rightarrow z$. The generator $G$ sharing with parameter $\theta_g$ across domains  maps multi-domain outputs $O$ into the latent feature $z$. 
We also define $m$ discriminators $\{D^i(z; \theta_d^i)\}_{i=1}^{m}$. 
Each discriminator $D^i(z): z \rightarrow [0,1]$ outputs a scalar representing the probability that $z$ comes from the $i^{th}$ domain rather than others.
For $D^i(z)$, we treat the correct label of an embedding in the $i^{th}$ domain, \textit{i.e.}, $O^i$, as \textit{True},  while others as \textit{False}. Then we divide multi-domain outputs into two groups, $\{O_{T}^i\}$ and $\{O_{F}^i\}$, which are defined as:
\begin{align}
    \{O_{T}^i\} \!=\! \{O^i\}; \{O_{F}^j\} \!=\! \{O^j| j\neq i \}; \bigcup_{j \neq i}\{O_{F}^j\} \!\cup\! \{O_{T}^i\} \!=\! O
\end{align}
The training goal of $D^i$ is to maximize the probability of assigning the correct label to features both from the $i^{th}$ domain and other domains, while the generator $G$ is trained against the discriminator to minimize $log(1-D(G(x)))$. In other words, it is a min-max game between the $i^{th}$ discriminator $D^i$ and generator $G$ with a value function $V$, formulated as:
\begin{equation}
\label{equ:minmax}
\begin{aligned}
\underset{G}{\textbf{min}} \; \underset{D^i}{\textbf{max}} \, V&(D^i, G) =\mathbb{E}_{x \sim \{O^i_{T}\}}[logD^i(G(x))] \\
& + \sum_{j \neq i}\mathbb{E}_{x\sim \{O^j_{F}\}}[log(1-D^i(G(x)))].
\end{aligned}
\end{equation}
During optimizing the min-max adversarial loss for $G$ and $D^i$, the distributions of model outputs from the $i^{th}$ domain and other domains become closer. 
After $G$ and all $D$ are well trained, $G$ will embed multi-domain model outputs into an invariant feature space, where each discriminator cannot figure out which domain the outputs of white-box models are from.

\subsection{Domain-agnostic Reverse Model}
Then, we use the domain-agnostic reverse classifier to classify the domain invariant features produced by the generator. 
We denote features $z$ produced by $G(O;\theta_g)$ as
\begin{equation}
  z = [G(O^1); G(O^2); ...; G(O^m)] \in \mathbb{R}^{m \times K \times d'}.
\end{equation} 
Where $d'$ is the number of feature dimensions. 
We define the domain-agnostic reverse classifier as $\Phi(z;\theta_c)$ parameterized by $\theta_c$. We obtain probability $p(z^i)$ for each possible model attribute  as:
\begin{equation}
  p(z^i) = softmax(\Phi(z^i)) = \frac{exp\{\Phi(z^i)\}}{\sum_{i=1}^m exp\{\Phi(z^j)\}}.
\end{equation}
The target is to minimize the cross entropy between the predicted $p(z^i)$ and ground-truth of model attribute values $y$:
\begin{equation}
  \label{equ:metace}
  \begin{aligned}
  \underset{\Phi}{\textbf{min}} \; \mathbb{E}_{z \sim G(O)} &\left[\sum_{i=1}^C -y^ilog(p(z^i))\right] = \\ 
 & \underset{\Phi}{\textbf{min}} \; \mathbb{E}_{z \sim G(O)} \left[-y^Tlog(p(z))\right].
  \end{aligned}
\end{equation}
At the inference phase, given the same queries as the white-box model, the outputs of a black-box model  from an unknown domain are fed into the generator $G$, and then the output of $G$ is fed into the  reverse classifier $\Phi$, achieving domain-agnostic prediction of black-box model attributes.
\begin{algorithm*}
  \caption{Training Strategy}
  
  \SetKw{Initialization}{Initialization}

    \KwIn{Batch size $b$, learning rate $\alpha$, $\beta$, multi-domain model outputs $O$, trade-off scalar $\lambda$}
    
    \KwOut{Generator $G$, meta-classifier $\Phi$, discriminators $\{D^i\}_{i=1}^m$}
    
    {\textbf{Initialize:} Initialize parameter $\theta_g$ of generator $G$, parameter $\theta_d^i$ of discriminators $\{D^i\}_{i=1}^m$ and parameter $\theta_c$ of domain-free meta-classifier $\Phi$ with normal distribution} \\
   
    \While {classifier $\Phi$ not converges}{
      Random sample $b$ samples $O_b^i$ from outputs $O^i$ in each domain\\
      \For {$i = 1, ..., m$}{
        Take samples in the $i^{th}$ domain as \textit{True} samples $X= O^i_b = \{x^1, x^2, ..., x^b\}$ \\
        \For {$j =1,...,m \;and\;j \neq i$}{ 
            Take samples in the $j^{th}$ domain as \textit{False} samples $\bar{X}_j= O^j_b = \{\bar{x}_j^1, \bar{x}_j^2, ..., \bar{x}_j^b\}$
        }
        Update the discriminator $D^i$ by gradient descent: \\
        $\theta_{d}^i := \theta_{d}^i - \alpha \nabla_{\theta_{d}^i}\left\{\sum_{k=1}^b \left[logD(G(x^k)\right] + \sum_{j\neq i}\left[\sum_{k=1}^blog(1-D(G(\bar{x}^k_j)))\right]\right\}$
      }
      
      Construct $X_{all} = X \cup \bar{X} = \{x_{all}^1, x_{all}^2, ..., x_{all}^{bm}\}$ and 
      $Z_{all} = G(X_{all}) = \{z_{all}^1, z_{all}^2, ..., z_{all}^{bm}\}$ \\
      Set the corresponding labels as $Y_{all} = \{y_{all}^1, y_{all}^2, ..., y_{all}^{bm}\}$ \\
      Calculate gradient of $\theta_c$ and $\theta_g$ by: \\
       $\bm{grad_{c}} = \nabla_{\theta_{c}}\sum_{k=1}^{bm}\left[-{y_{all}^k}^Tlog(p(z_{all}^k))\right]$ \\
      $\bm{grad_{g}} = \nabla_{\theta_{g}, \theta_{c}}\left\{\sum_{j\neq i}\sum_{k=1}^b\left[log(1-D(G(\bar{x}^k_j)))\right] - \lambda\sum_{k=1}^{bm}\left[{y_{all}^k}^Tlog(p(z_{all}^k))\right]\right\}$ \\
    Update the classifier $\Phi$ and generator $G$ together: \\
      $\theta_c := \theta_c - \beta \cdot \bm{grad_{c}}$ and 
      $\theta_g := \theta_g - \alpha \cdot \bm{grad_{g}}$
    } 
  \label{alg:training}
\end{algorithm*}
\subsection{Overall Model and Training Strategy}
After introducing all the components, we give the final loss function based on Eq. \ref{equ:minmax} and \ref{equ:metace} as:
\begin{equation}
  \label{equ:final}
  \begin{aligned}
    \underset{G, \Phi}{\textbf{min}} \; \underset{D^i, 1 \leq i \leq m}{\textbf{max}} \, V&(D^i, G)  = \mathbb{E}_{x \sim \{O^i_{T}\}}\left[logD^i(G(x))\right] \! \\
    & + \! \sum\nolimits_{j \neq i}\mathbb{E}_{x\sim \{O^j_{F}\}}\left[log(1-D^i(G(x)))\right]  \\
    & + \lambda\  \mathbb{E}_{z \sim G(O)} \ \left[-y^Tlog(p(z))\right]. \nonumber
  \end{aligned}
\end{equation}
where $\lambda$ is a trade-off parameter.

The training strategy is as follows: we first optimize all discriminators $D^i$, and then jointly optimize the generator and the domain-agnostic reverse classifier. We repeat the above processes until the algorithm converges.
 The proposed training strategy is represented in Algorithm \ref{alg:training}.

\section{Experiments}

\begin{table}[t]
\renewcommand\arraystretch{0.75}
\centering
\caption{Attributes and the corresponding values.}
\begin{tabular}{ll}\\
\toprule[1.3pt]
\textbf{Attribute}  & \textbf{Values} \\ \toprule [1.3pt]
\#Activation & ReLU, PReLU, ELU, Tanh \\  \midrule
\#Dropout & Yes, No  \\  \midrule
\#Max pooling & Yes, No  \\  \midrule
\#Batchnorm & Yes, No   \\  \midrule
\#Kernel size & 3, 5   \\  \midrule
\#Conv layers & 2, 3, 4  \\  \midrule
\#FC layers & 2, 3, 4   \\  \midrule
\#Optimizer & SGD, ADAM, RMSprop   \\  \midrule
\#Batch size & 32, 64, 128  \\  \midrule[1.3pt]
\end{tabular}
\label{table:attr}
\end{table}

\begin{table*}[t]
  \renewcommand\arraystretch{1}
  \caption{Model attribute classification accuracy (\%) on PACS-modelset. \textcolor{red}{\textbf{Red}} and \textcolor{blue}{\underline{blue}}  indicate the best and second best performance, respectively.}
  \label{table:result_PACS}
  \centering
  \setlength\tabcolsep{5pt}
  \begin{tabular}{c|cccccccccccc}
  \toprule[1.3pt]
  \multirow{2}{*}{} & \multirow{2}{*}{Method} & \multicolumn{9}{c}{Attributes} & \multirow{2}{*}{Avg}  \\
  \cline{3-11}
   & & \#act & \#drop & \#pool & \#ks & \#conv & \#fc & \#opt& \#bs & \#bn \\
   \toprule[1.3pt]
  &Random &25.00 &50.00 &50.00 &50.00 &33.33&33.33 &33.33 &33.33 &50.00 &39.81 \\
  \toprule[1.3pt]
  \multirow{8}{*}{P}&SVM &37.80 &50.30 &54.80 &53.60 &34.00 &36.60 &37.00 &\color{blue}\underline{45.70} &58.80 &45.40 \\
  &KENNEN* &39.07 &50.68 &59.42 &61.31&\color{blue}\underline{36.18} &39.33 &37.88 &44.16 &59.74 &47.53\\
  &SelfReg &25.58 &52.26 &54.98 &50.18 &34.12 &35.25 &34.61 &33.78 &50.76 &41.28 \\
  &MixStyle &\color{blue}\underline{39.63} &53.23 &\color{blue}\underline{61.83} &59.44 &35.66 &38.75 &37.89 &43.75 &57.09 &47.47 \\
  &MMD &38.88 &\color{blue}\underline{54.70} &60.46 &56.54 &35.38 &36.66 &35.66 &40.50 &61.04 &46.65 \\
  
  &SD &38.70 &51.06 &58.86 &\color{blue}\underline{62.21} &35.84 &\color{blue}\underline{40.05} &\color{blue}\underline{39.23} &44.34 &\color{blue}\underline{62.12} &\color{blue}\underline{48.04} \\
  &\textbf{DREAM} &\color{red}\textbf{43.84} &\color{red}\textbf{59.19} &\color{red}\textbf{66.09} &\color{red}\textbf{64.24} &\color{red}\textbf{39.59} &\color{red}\textbf{42.04} & \color{red}\textbf{40.49} &\color{red}\textbf{47.83} &\color{red}\textbf{68.12} &\color{red}\textbf{52.38} \\
  \midrule[1.3pt]
  \multirow{8}{*}{C} &SVM &25.80 &49.20 &50.70 &55.80 &\color{blue}\underline{37.20} &38.10 &30.80 &\color{blue}\underline{42.30} & \color{blue}\underline{65.30} &43.91 \\
  &KENNEN* &32.99 &52.50 &54.23 &56.57 &37.19 &\color{blue}\underline{40.53} &33.47 &37.17 &\color{red}\textbf{68.39} &45.89\\
  &SelfReg &25.97 &51.42 &\color{blue}\underline{56.20} &50.03 &35.04 &35.52 &\color{blue}\underline{36.09} &35.58 &56.17 &42.44 \\
  &MixStyle &32.10 &50.76 &55.44 &54.18 &36.18 &37.87 &34.65 &38.69 &60.26 &44.46 \\
  &MMD &29.56 &53.02 &54.70 &53.82 &35.38 &36.36 &35.98 &37.24 &57.58 &43.75 \\
  
  &SD &\color{blue}\underline{33.52} &\color{blue}\underline{54.06} &54.12 &\color{blue}\underline{56.69} &36.84 &\color{red}\textbf{41.02} &35.61 &36.12 &65.12 &\color{blue}\underline{45.90} \\
  
  &\textbf{DREAM} &\color{red}\textbf{37.53} &\color{red}\textbf{55.89} &\color{red}\textbf{61.18} &\color{red}\textbf{57.32} &\color{red}\textbf{38.58} &39.60 &\color{red}\textbf{38.32} &\color{red}\textbf{45.01} &65.16 &\color{red}\textbf{48.73} \\
  \midrule[1.3pt]
  
  \multirow{8}{*}{S}&SVM &23.80 &47.60 &47.40 &45.80 &33.80 &34.50 &31.80 &34.30 &53.10 &39.12 \\
  &KENNEN* &34.64 &50.10 &53.07 &52.01 &34.61 &37.11 &35.78 &\color{blue}\underline{37.04} &55.27 &43.29\\
  &SelfReg &27.07 &\color{blue}\underline{54.32} &51.39 &53.07 &36.99 &36.82 &35.47 &34.17 &\color{blue}\underline{61.80} &43.46 \\
  &MixStyle &\color{blue}\underline{37.78} &51.71 &54.16 &\color{blue}\underline{53.60} &34.53 &36.16 &\color{blue}\underline{36.36} &36.02 &59.42 &44.42 \\
  &MMD &31.96 &52.94 &56.84 &52.78 &\color{blue}\underline{38.18} &\color{blue}\underline{38.20} &36.20 &35.92 &57.56 &\color{blue}\underline{44.51} \\

  &SD &34.82 &52.51 &\color{blue}\underline{56.89} &51.21 &34.23 &38.12 &35.91 &36.72 &54.23 &43.85 \\
  
  &\textbf{DREAM}
  &\color{red}\textbf{39.71} &\color{red}\textbf{57.74} &\color{red}\textbf{64.73} &\color{red}\textbf{60.79} &\color{red}\textbf{40.79} &\color{red}\textbf{40.14} &\color{red}\textbf{43.54} &\color{red}\textbf{43.80} &\color{red}\textbf{72.51} &\color{red}\textbf{51.53} \\
  \bottomrule[1.3pt]
\end{tabular}
\end{table*}

\label{others}

\subsection{Dataset Construction}
\label{subsec:dataset_construction}
Following \cite{oh2018towards2}, we train a number of models which are constructed by enumerating all possible attribute values.
The details of the attributes and their values are shown in Table \ref{table:attr}.
The number of models with all possible combinations of the attributes is $5,184$. 
We also initialize each model with random seeds from $0$ and $999$, yielding 5,184,000 unique white-box models. 
For each domain, we randomly select and train 10,000 white-box models from 5,184,000 models. Then we sample 5,000, 1,000, 1,000 from 10,000 white-box models as the training modelset, validation modelset, and testing modelset. 
We use PACS-modelset and MEDU-modelset to evaluate our method.

\textbf{PACS-modelset}. 
PACS is an image dataset that has been widely used for OOD learning \cite{li2017deeper}. 
In this experiment, we use it for evaluating our domain-agnostic black-box model attribute inference framework DREAM. We utilize three domains, including Photo (1,670 images), Cartoon (2,344 images), and Sketch (3,929 images), to construct our dataset. In our dataset, each domain contains 7 categories. 
For each domain, we train 10,000 models and we combine them as PACS-modelset (30,000 models in total).

\textbf{MEDU-modelset}.
MEDU is a set of hand-written digit recognition datasets, with 4 domains collected from MNIST \cite{lecun_mnist_1998}, USPS \cite{hull_database_1994}, DIDA \cite{DIDA2020} and EMNIST \cite{cohen_emnist_2017}. Each domain contains different styles of hand-written digit from 0 to 9. We train 40,000 models as MEDU-modelset and each domain contains 10,000 models. 

We construct the two modelsets (PACS-modelset and MEDU-modelset) by enumerating combinations of attribute values. The attribute is shown in Table \ref{table:attr}. The architecture of each model in modelsets follows the scheme: $N_c$ convolution layers, $N_f$ fully-connected layers, and a linear classifier. Each convolution layer contains a $k \times k$ convolution, an optional batch normalization, an optional max-pooling, and a non-linear activation function in sequence, where $k$ is the kernel size. Each fully-connected layer consists of a linear transformation, a non-linear activation, and an optional dropout in sequence. We set the dropout ratio to $0.1$ in our experiments. When training white-box models, optimizers are selected from $\{$SGD, ADAM, RMSprop$\}$ with a batch size $32$, $64$ or $128$, respectively. \textcolor{black}{The statistics of PACS-modelset and MEDU-modelset are listed in Table \ref{PACS-P} to \ref{PACS-S} and Table \ref{MEDU-M} to \ref{MEDU-U}, respectively}

\begin{table*}[t]
\renewcommand\arraystretch{1}
  \caption{Model attribute classification accuracy (\%) on MEDU-modelset. \textcolor{red}{\textbf{Red}} and \textcolor{blue}{\underline{blue}}  indicate the best and second best performance, respectively.}
  \label{table:result_MEDU}
  \centering
  \setlength\tabcolsep{5pt}
  \begin{tabular}{c|cccccccccccc}
  \toprule[1.3pt]
  \multirow{2}{*}{} & \multirow{2}{*}{Method} & \multicolumn{9}{c}{Attributes} & \multirow{2}{*}{Avg}  \\
  \cline{3-11}
   & & \#act & \#drop & \#pool & \#ks & \#conv & \#fc & \#opt& \#bs & \#bn \\
     \toprule[1.3pt]
  &Random &25.00 &50.00 &50.00 &50.00 &33.33&33.33 &33.33 &33.33 &50.00 &39.81 \\
  \toprule[1.3pt]
  \multirow{8}{*}{M} &SVM &45.60 &49.40 &62.90 &\color{red}\textbf{59.20} &38.80 &\color{red}\textbf{40.10} &35.50 &35.00 &75.30 &49.09 \\
  &KENNEN* &\color{red}\textbf{51.18} &50.67 &62.99 &57.36 &38.32 &35.84 &41.57 &35.75 &77.87 &50.17\\
  &SelfReg &28.00 &53.57 &53.43 &50.78 &35.97 &36.39 &35.98 &36.23 &53.96 &42.70 \\
  &MixStyle &50.27 &51.72 &62.66 &57.32 &37.88 &36.34 &\color{blue}\underline{43.11} &\color{blue}\underline{38.00} &\color{red}\textbf{82.61} &51.10 \\
  &MMD &44.57 &\color{blue}\underline{59.67} &\color{red}\textbf{66.37} &57.27 &\color{blue}\underline{39.63} &37.27 &42.10 &37.60 &81.37 &\color{blue}\underline{51.76} \\
  & SD & 49.60 & 49.40& 62.40& 52.30& 37.10& 36.70& 38.90& 35.30& 81.50& 49.24\\
  
  &\textbf{DREAM} &\color{blue}\underline{51.01} &\color{red}\textbf{62.32} &\color{blue}\underline{64.28} &\color{blue}\underline{58.39} &\color{red}\textbf{40.96} &\color{blue}\underline{38.11} &\color{red}\textbf{45.37} &\color{red}\textbf{38.96} &\color{blue}\underline{81.99} &\color{red}\textbf{53.49} \\
  \midrule[1.3pt]
  \multirow{8}{*}{E}&SVM &40.00 &48.70 &\color{blue}\underline{69.20} &51.60 &40.20 &36.90 &35.80 &30.10 &79.90 &48.04 \\
  &KENNEN* &\color{red}\textbf{45.66} &51.01 &65.26 &53.25 &40.28 &36.35 &41.96 &36.16 &81.30 &50.14\\
  &SelfReg &27.29 &52.83 &53.32 &52.85 &33.68 &35.05 &35.26 &35.32 &53.74 &42.15 \\
  &MixStyle &43.68 &51.35 &67.87 &57.15 &42.50 &\color{blue}\underline{39.30} &\color{blue}\underline{42.10} &38.79 &82.46 &51.69 \\
  &MMD &42.03 &\color{blue}\underline{58.43} &66.27 &\color{red}\textbf{60.80} &40.80 &38.67 &40.00 &\color{blue}\underline{39.97} &84.00 &\color{blue}\underline{52.33} \\
  & SD & 43.60& 48.90& 59.20& 60.10& \color{red}\textbf{44.50}& 35.10& 43.60& 33.80& \color{blue}\underline{88.80}& 50.84\\
  &\textbf{DREAM} &\color{blue}\underline{45.55} &\color{red}\textbf{64.98} &\color{red}\textbf{74.16} &\color{blue}\underline{60.71} &44.45&\color{red}\textbf{42.45} &\color{red}\textbf{47.37} &\color{red}\textbf{41.03} &\color{red}\textbf{91.00} &\color{red}\textbf{56.86} \\
  \midrule[1.3pt]
  \multirow{8}{*}{D}&SVM &45.00 &47.80 &54.60 &45.50 &29.40 &37.60 &\color{red}\textbf{43.30} &36.50 &\color{red}\textbf{63.70} &44.82 \\
  &KENNEN* &42.73 &52.06 &55.27 &52.02 &34.89 & 38.90 &38.98 &36.27 &54.97 &45.12\\
  &SelfReg &26.31 &54.29 &53.23 &52.33 &34.96 &35.72 &36.49 &35.39 &59.11 &43.09 \\
  &MixStyle &\color{blue}\underline{45.26} &52.32 &55.91 &51.39 &34.22 &38.70 &38.31 &\color{blue}\underline{38.03} &57.44 &45.73 \\
  &MMD &39.00 &\color{blue}\underline{59.20} &\color{red}\textbf{59.63} &\color{blue}\underline{55.93} &\color{blue}\underline{35.93} &38.33 &37.93 &37.50 &54.40 &46.43 \\
  & SD & 46.50& 51.80& 52.70& 51.90& 34.90& \color{red}\textbf{45.30}& \color{blue}\underline{42.70}& 36.80& 59.50&  \color{blue}\underline{46.92}\\
  &\textbf{DREAM} &\color{red}\textbf{49.63} &\color{red}\textbf{64.50} &\color{blue}\underline{59.30} &\color{red}\textbf{57.13} &\color{red}\textbf{39.52} &\color{blue}\underline{44.59} &42.09 &\color{red}\textbf{40.19} &\color{blue}\underline{59.68} &\color{red}\textbf{50.74} \\
  \midrule[1.3pt]

  \multirow{8}{*}{U}&SVM &\color{red}\textbf{43.40} &50.50 &47.60 &52.50 &30.30 &32.30 &\color{red}\textbf{41.00} &36.60 &49.40 &42.62 \\
  &KENNEN* &\color{blue}\underline{43.38} &50.88 &51.41 &53.19 &36.35&35.59 &36.66 &34.56 &55.62 &44.18\\
  &SelfReg &26.81 &52.16 &55.46 &52.47 &36.18 &\color{blue}\underline{36.43} &36.53 &35.90 &55.34 &43.03 \\
  &MixStyle &41.05 &53.80 &50.49 &52.93 &35.26 &33.68 &36.92 &34.75 &59.34 &44.25 \\
  &MMD &39.33 &\color{blue}\underline{55.87} &52.67 &\color{blue}\underline{53.23} &\color{red}\textbf{39.20} &34.33 &35.90 &\color{blue}\underline{36.90} &60.73 &\color{blue}\underline{45.35} \\
  & SD & 41.90& 50.40& \color{blue}\underline{58.00}& 52.30& 33.10& 35.30& 36.60& 34.30 & \color{blue}\underline{61.60}& 44.83 \\
  &\textbf{DREAM} &42.34 &\color{red}\textbf{58.72} &\color{red}\textbf{58.58} &\color{red}\textbf{54.41} &\color{blue}\underline{37.90} &\color{red}\textbf{37.81} &\color{blue}\underline{40.42} &\color{red}\textbf{38.36} &\color{red}\textbf{63.39} &\color{red}\textbf{47.99} \\
  
  \bottomrule[1.3pt]
\end{tabular}
\end{table*}

\subsection{Implementation Details of DREAM}
\label{app:impl_details}
In the experiment, we set the number of queries $N$ to 100. 
We use Adam \cite{kingma2014adam} as the optimizer, where the learning rate $\alpha$  is set to $10^{-5}$  for the generator and discriminators, and the learning rate $\beta$ is set to $10^{-4}$ for the reverse model. 
The batch size $b$ is set as 100. 
The trade-off parameter $\lambda$ is tuned from $\{0.001, 0.01, 0.1, 1, 10\}$ based on the validation set. 

In addition, the multi-discriminator GAN is composed of a generator and multiple discriminators. The generator consists of two linear layers with ReLU activation. The dimension of the input layer of the generator is determined by the query number $N$ and class category number $C$.  In the experiment of recognizing handwritten digits, the input dimension is 1000 ($N=100$, $C=10$). In the case of the PACS dataset, the input dimension is 700 ($N=100$, $C=7$), and the output dimension of the successive two layers is respectively 500 and 128. Each discriminator consists of three linear layers, with ReLU activation and a final Sigmoid activation. The output dimension of layers is 512, 256, 1 respectively. All experiments are conducted on 4 NVIDIA RTX 3090 GPUs, PyTorch 1.11.0 platform \cite{paszke2019pytorch}.

\subsection{Baselines}
We compare our DREAM with 7 baselines including Random choice, SVM, KENNEN \cite{oh2018towards2}, SelfReg \cite{kim2021selfreg}, MixStyle \cite{zhou2021mixstyle}, MMD \cite{li_domain_2018} and SD \cite{pezeshki2021gradient}. To compare fairly, we select a variant of KENNEN (denoted as KENNEN*) taking fixed queries as input, which is the same as ours. We take SVM as a baseline without considering different domain outputs. Moreover, we also take three typical OOD generalization methods, SelfReg, MixStyle, MMD and SD as baselines to verify the effect of our framework for learning domain invariant features. 
We adopt the following training strategies for baselines:

\textcolor{black}{
\textbf{SVM.} We directly input the multi-domain probability outputs into SVM classifiers to predict attributes. Note that the SVM does not differentiate between outputs from different domains
}

\textcolor{black}{
\textbf{KENNEN.} We embed the multi-domain probability outputs into feature space and apply classifiers to predict attributes. Note that the KENNEN does not differentiate between outputs from different domains. For a fair comparison, the network structure is the same as our domain-agnostic reverse model.
}

\textcolor{black}{
\textbf{SelfReg, MixStyle, MMD and SD.} We embed the multi-domain probability outputs into feature space and apply the domain generalization method, SelfReg, MixStyle, MMD and SD, respectively to learn invariant features. Then the invariant features are input to classifiers to predict attributes. For a fair comparison, the network structure of classifiers is the same as our domain-agnostic reverse model.
}

We adopt ``leave-one-domain-out" scheme to split the source and target domains. For each dataset, we in turn take one domain as the target domain and the rest  as source domains. We run the experiment 10 trials and report the average result.

\begin{table*}[h]
  \caption{Accuracy of model extraction using different extraction model structures. The extraction model structures are "same to the victim", "random", or "inferred by DREAM".
}
  \label{table:me}
  \centering
  \setlength\tabcolsep{5pt}
  \begin{threeparttable}
  \begin{tabular}{ccccc}
  \toprule[1.3pt]
  \multirow{2}{*}{Dataset} & 
  \multirow{2}{*}{Victim Model Accuracy} &
  \multicolumn{3}{c}{Extraction Model Structure} \\  
  \cline{3-5}
   & & \#Same to Victim & \#Random & \# Inferred by DREAM  \\
   \toprule[1.3pt]
 
  \multicolumn{1}{c}{MNIST} &86.43\% & $\mathbf{68.46\%_{(0.79\times)}}$
  &$45.88\%_{(0.53\times)}$
  &$62.81\%_{(0.73\times)}$  \\
  \bottomrule[1.3pt]
\end{tabular}
\end{threeparttable}
\end{table*}

\begin{figure*}[t]
	\centering
	\setcounter {subfigure} {0} a){
    	\begin{minipage}{0.22\linewidth}
    		\centering
    		\includegraphics[width=1\linewidth]{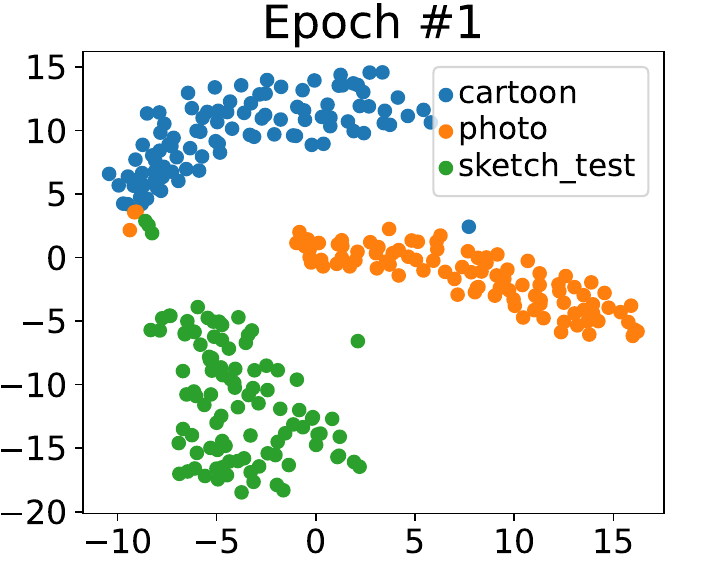}
    		\label{fig:param_id2_S}
    	\end{minipage}
    	\begin{minipage}{0.22\linewidth}
    		\centering
    		\includegraphics[width=1\linewidth]{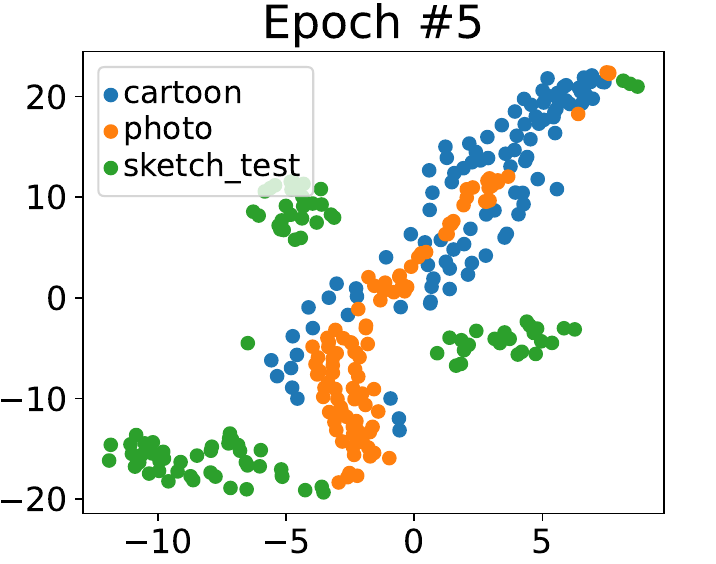}
    		\label{fig:param_id2_P}
    	\end{minipage}
    	\begin{minipage}{0.22\linewidth}
    		\centering
    		\includegraphics[width=1\linewidth]{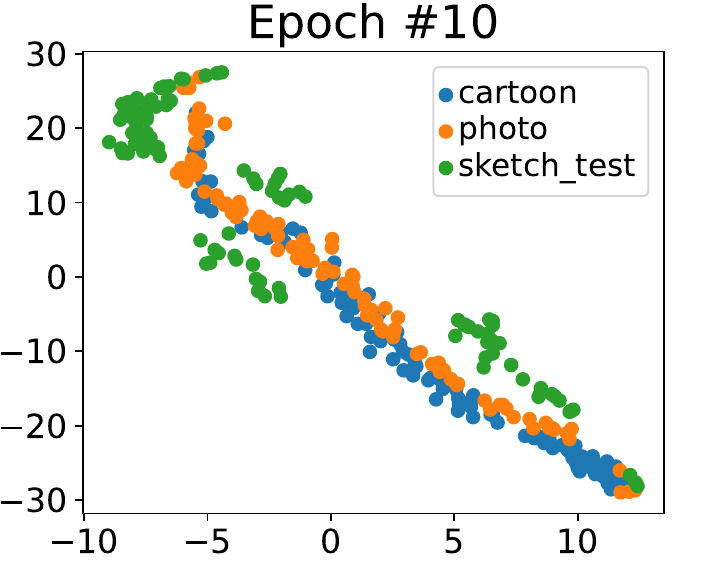}
    		\label{fig:param_id2_C}
    	\end{minipage}
    	\begin{minipage}{0.22\linewidth}
    		\centering
    		\includegraphics[width=1\linewidth]{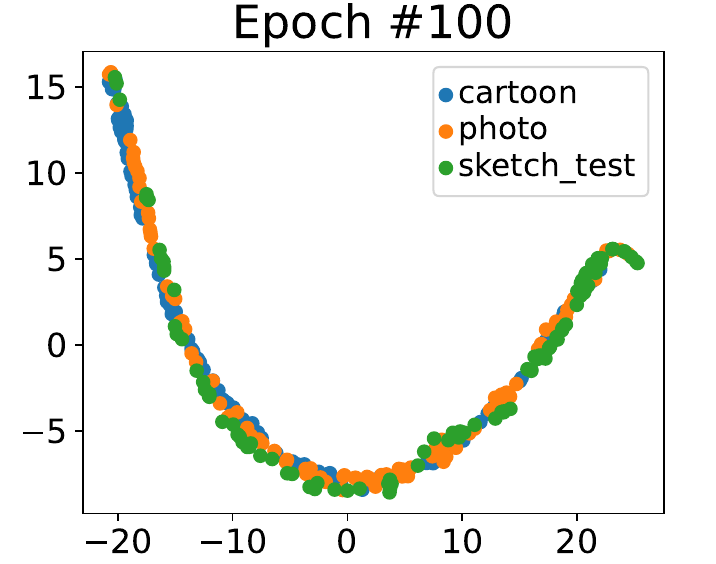}
    		\label{fig:param_id2_S}
    	\end{minipage}
	}
	\\
	\setcounter {subfigure} {0} b){
	    \begin{minipage}{0.22\linewidth}
		\centering
		\includegraphics[width=1\linewidth]{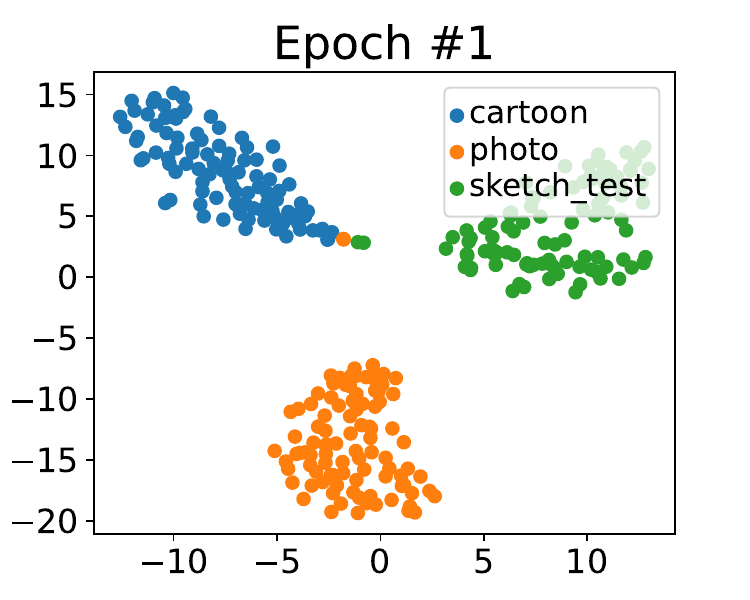}
		\label{fig:param_id2_S}
	\end{minipage}
	\begin{minipage}{0.22\linewidth}
		\centering
		\includegraphics[width=1\linewidth]{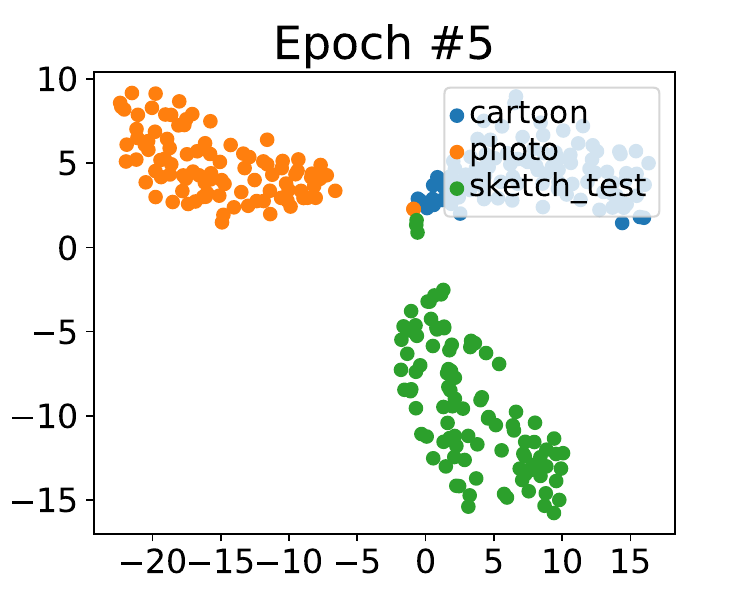}
		\label{fig:param_id2_P}
	\end{minipage}
	\begin{minipage}{0.22\linewidth}
		\centering
		\includegraphics[width=1\linewidth]{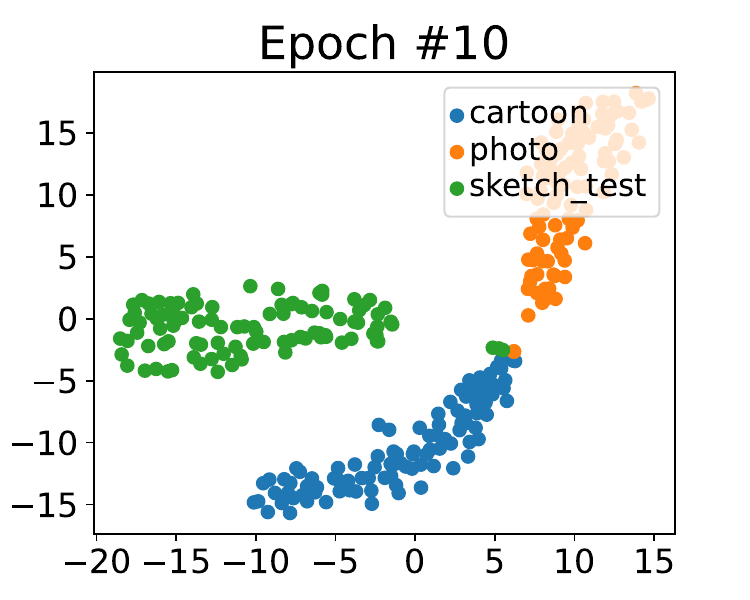}
		\label{fig:param_id2_C}
	\end{minipage}
	\begin{minipage}{0.22\linewidth}
		\centering
		\includegraphics[width=1\linewidth]{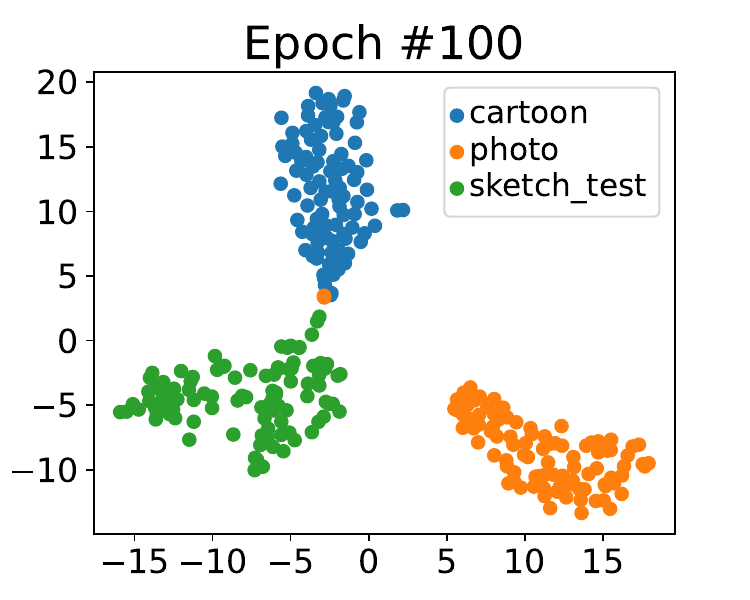}
		\label{fig:param_id2_S}
	\end{minipage}
	}
	\\
	\setcounter {subfigure} {0} c){
	    \begin{minipage}{0.22\linewidth}
    		\centering
    		\includegraphics[width=1\linewidth]{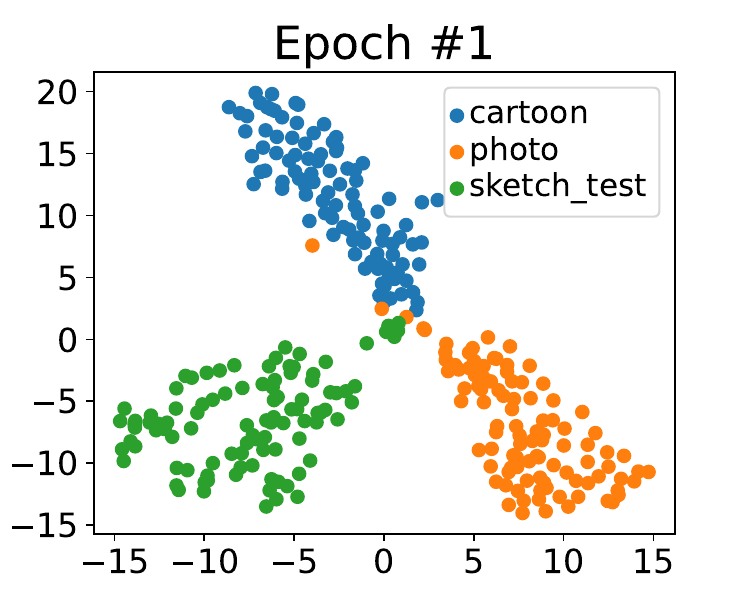}
    		\label{fig:param_id2_S}
    	\end{minipage}
    	\begin{minipage}{0.22\linewidth}
    		\centering
    		\includegraphics[width=1\linewidth]{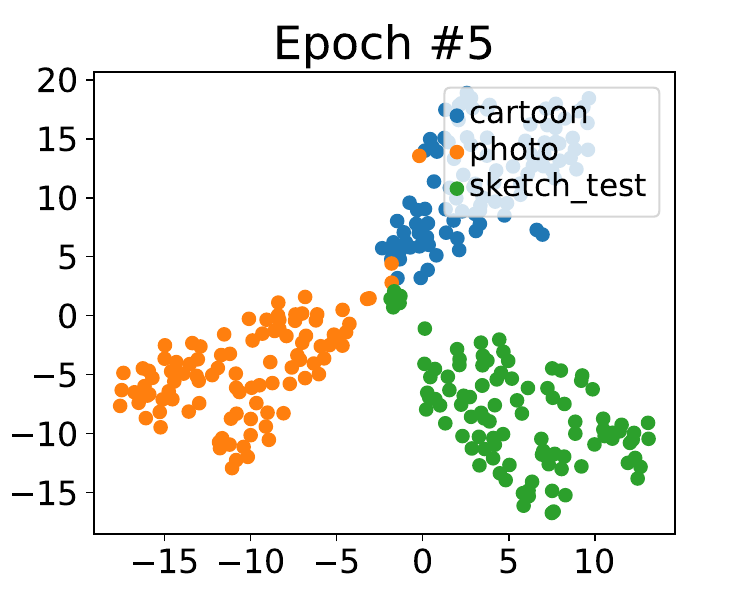}
    		\label{fig:param_id2_P}
    	\end{minipage}
    	\begin{minipage}{0.22\linewidth}
    		\centering
    		\includegraphics[width=1\linewidth]{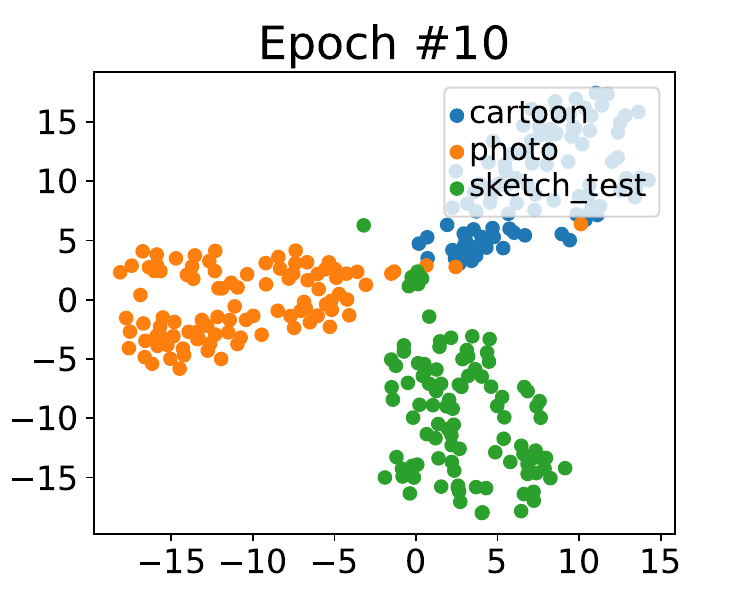}
    		\label{fig:param_id2_C}
    	\end{minipage}
    	\begin{minipage}{0.22\linewidth}
    		\centering
    		\includegraphics[width=1\linewidth]{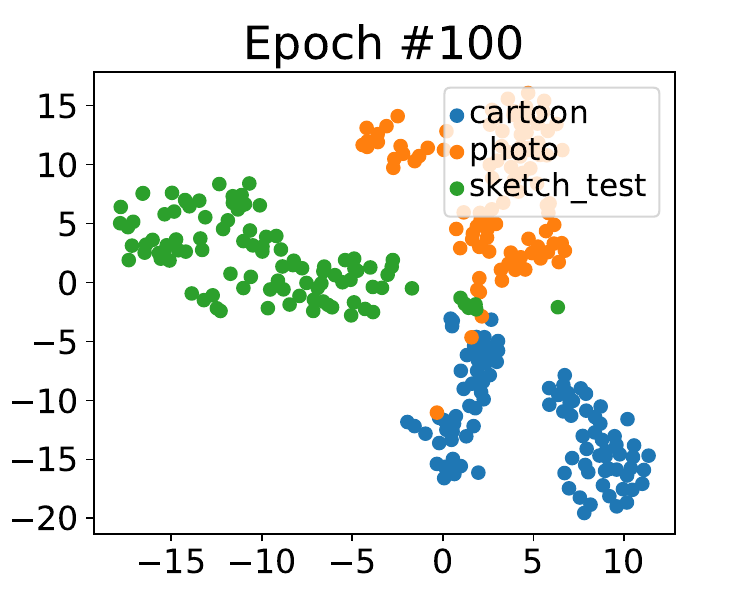}
    		\label{fig:param_id2_S}
    	\end{minipage}
	}
	\\
	\setcounter {subfigure} {0} d){
	    \begin{minipage}{0.22\linewidth}
		\centering
		\includegraphics[width=1\linewidth]{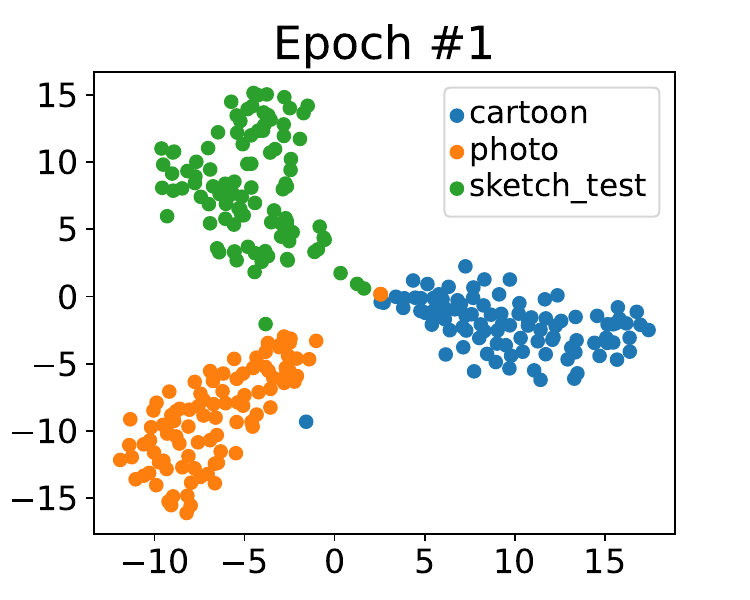}
		\label{fig:param_id2_S}
    	\end{minipage}
    	\begin{minipage}{0.22\linewidth}
    		\centering
    		\includegraphics[width=1\linewidth]{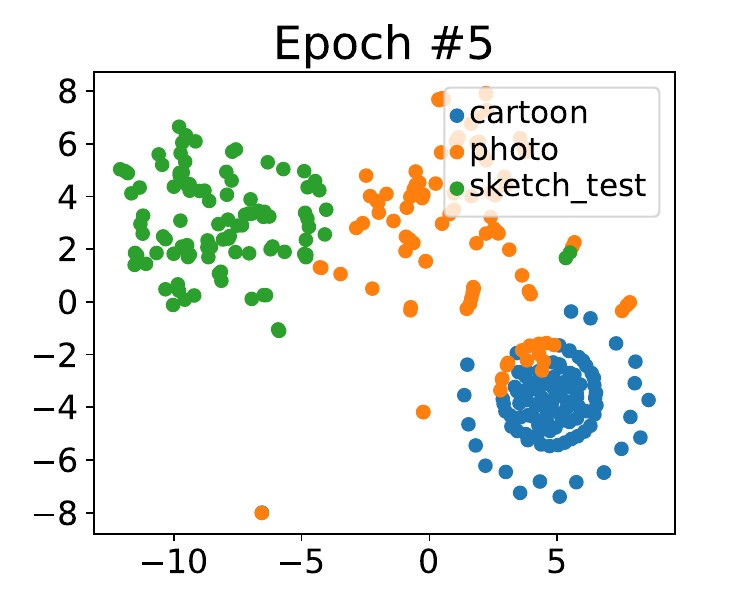}
    		\label{fig:param_id2_P}
    	\end{minipage}
    	\begin{minipage}{0.22\linewidth}
    		\centering
    		\includegraphics[width=1\linewidth]{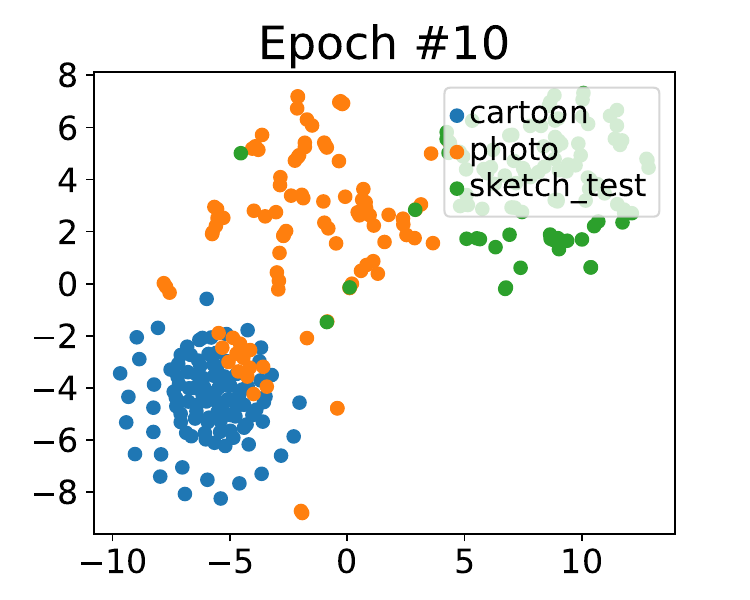}
    		\label{fig:param_id2_C}
    	\end{minipage}
    	\begin{minipage}{0.22\linewidth}
    		\centering
    		\includegraphics[width=1\linewidth]{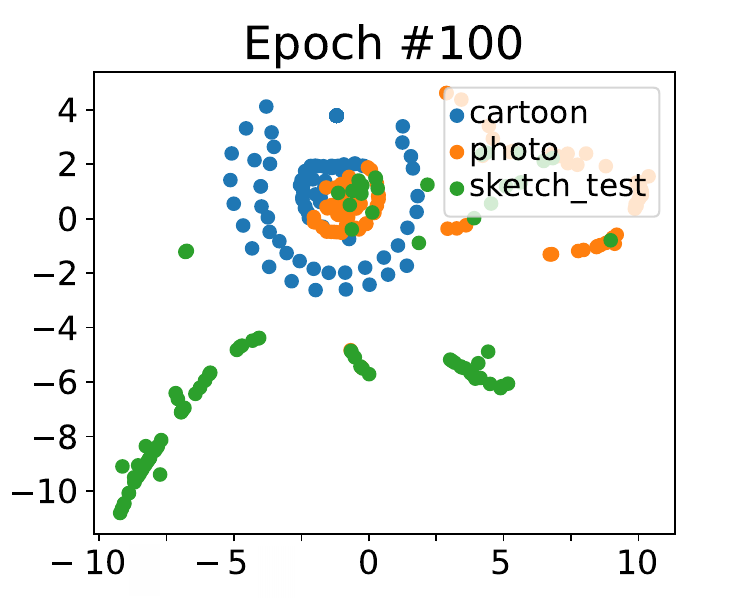}
    		\label{fig:param_id2_S}
    	\end{minipage}
	}
	\caption{T-SNE visualization of features of different domains produced by a) DREAM, b) MMD, c) MisStyle and d) SelfReg on PACS-modelset.}
	\label{fig:tsne_DREAM} 
\end{figure*}

\subsection{Comparison with Baselines}
\label{subsec:exp_result}
Table \ref{table:result_PACS} and \ref{table:result_MEDU} report the overall performance of different methods on the PACS-modelset and MEDU-modelset, respectively. The leftmost column in each table indicates the target domain (the rest ones are source domains). 
The performance achieved by our proposed DREAM is better than that of all baselines in terms of the average result of model attributes.
For individual attribute, our method outperforms other methods in most cases.
Our method is better than KENNEN, which illustrates our method benefits from learning domain invariant features and learning the domain-agnostic reverse model. 
Moreover, our method achieves better performance than the three OOD learning methods. 
SelfReg draws samples of similar categories between all domains closer and samples of different categories farther. It may be difficult to simultaneously draw samples of 9 attributes closer or farther. MixStyle learns image style information and performs style mixing. However, it might not capture styles of probability data well.
What is more, we observe that DREAM cannot outperform other baselines in some cases. The reasons might be: 1) DREAM vs. OOD learning baselines. As we have mentioned, the OOD learning methods aim to learn a domain invariant space from different domains. Once features of different domains are excessively pulled close, the classification accuracy would be influenced. Thus, the trade-off between invariant space learning and classification learning is vital for the performance of reverse engineering. Moreover, the trade-off for each attribute is not identical. Taking MMD in Table \ref{table:result_MEDU} E as an example, the best trade-off hyperparameter conduces to predict attribute \#ks better than other attributes.  Similarly, the best trade-off hyperparameter of DREAM conduces to better predict attributes except for \#ks. 2) DREAM vs. KENNEN and SVM. Our proposed DREAM has stronger ability to fit complicated data, while SVM and KENNEN (only a shallow MLP) are weaker in the scenario of complicated data. 

\subsection{Application Value of Model Reverse Engineering}
To show the application value of our method, we conduct a model extraction experiment. We adopt a popular model extraction method MAZE \cite{kariyappa_maze_2021}, where model structures are ``same to the victim", ``random", or ``inferred by DREAM".
As shown in Table \ref{table:me}, we obtain $62.81\%$ extraction accuracy when we use the architecture inferred by DREAM, which outperforms random architecture and is close to the architecture of "same to the victim". \textcolor{black}{This is because the extraction performance increases as the structure of the extracted model gets closer to that of the victim \cite{sha2023can}. The closer the extracted model is to the victim model in terms of structure and complexity, the more likely it is to capture the same patterns and information as the victim model.}


\begin{table*}[h]
  \caption{Model attribute classification accuracy (\%) on S of PACS-modelset. \textcolor{red}{\textbf{Red}} and \textcolor{blue}{\underline{blue}}  indicate the best and second best performance, respectively. \textbf{DREAM*} is the result of domain shift
    scenario, trained with only five classes (except dog and elephant), while the black-box model is trained by whole seven classes in PACS dataset.}
  \label{table:ccs}
  \centering
  \setlength\tabcolsep{5pt}
  \begin{threeparttable}
  \begin{tabular}{c|cccccccccccc}
  \toprule[1.3pt]
  \multirow{2}{*}{} & \multirow{2}{*}{Method} & \multicolumn{9}{c}{Attributes} & \multirow{2}{*}{Avg}  \\
  \cline{3-11}
   & & \#act & \#drop & \#pool & \#ks & \#conv & \#fc & \#opt& \#bs & \#bn \\
   \toprule[1.3pt]
  &Random &25.00 &50.00 &50.00 &50.00 &33.33&33.33 &33.33 &33.33 &50.00 &39.81 \\
  \toprule[1.3pt]
  \multirow{8}{*}{S}&SVM &23.80 &47.60 &47.40 &45.80 &33.80 &34.50 &31.80 &34.30 &53.10 &39.12 \\
  &KENNEN* &34.64 &50.10 &53.07 &52.01 &34.61 &37.11 &35.78 &37.04 &55.27 &43.29\\
  &SelfReg &27.07 &54.32 &51.39 &53.07 &36.99 &36.82 &35.47 &34.17 &61.80 &43.46 \\
  &MixStyle &37.78 &51.71 &54.16 &53.60 &34.53 &36.16 &36.36 &36.02 &59.42 &44.42 \\
  &MMD &31.96 &52.94 &56.84 &52.78 &38.18 &38.20 &36.20 &35.92 &57.56 &44.51 \\
  &SD &34.82 &52.51 &56.89 &51.21 &34.23 &38.12 &35.91 &36.72 &54.23 &43.85 \\
  &\textbf{DREAM} 
  &\color{blue}\underline{39.71} &\color{red}\textbf{57.74} &\color{red}\textbf{64.73} &\color{red}\textbf{60.79} &\color{red}\textbf{40.79} &\color{red}\textbf{40.14} &\color{red}\textbf{43.54} &\color{red}\textbf{43.80} &\color{blue}\underline{72.51} &\color{red}\textbf{51.53} \\

  &\textbf{DREAM*}
&\color{red}\textbf{42.24} &\color{blue}\underline{55.68} &\color{blue}\underline{61.82} &\color{blue}\underline{58.34} &\color{blue}\underline{39.55} &\color{blue}\underline{38.39} &\color{blue}\underline{38.51} &\color{blue}\underline{41.39} &\color{red}\textbf{74.39} &\color{blue}\underline{50.03} \\
  \bottomrule[1.3pt]
\end{tabular}
\end{threeparttable}
\end{table*}

\begin{table*}
  \caption{Model attribute classification accuracy (\%) on P of PACS-modelset using different training and testing attribute combinations. \textcolor{red}{\textbf{Red}} and \textcolor{blue}{\underline{blue}}  indicate the best and second best performance, respectively.}
  \label{table:diff_arc}
  \centering
  \setlength\tabcolsep{5pt}
  \begin{threeparttable}
  \begin{tabular}{c|cccccccccccc}
  \toprule[1.3pt]
  \multirow{2}{*}{} & \multirow{2}{*}{Method} & \multicolumn{9}{c}{Attributes} & \multirow{2}{*}{Avg}  \\
  \cline{3-11}
   & & \#act & \#drop & \#pool & \#ks & \#conv & \#fc & \#opt& \#bs & \#bn \\
   \toprule[1.3pt]
  &Random &25.00 &50.00 &50.00 &50.00 &33.33&33.33 &33.33 &33.33 &50.00 &39.81 \\
  \toprule[1.3pt]
  \multirow{7}{*}{P}&SVM &34.20 &51.70 &48.50 &56.10 &35.70 &36.50 &37.60 &40.50 &64.60 &45.04 \\
  &KENNEN* &37.36 &53.12 &57.79 &59.66 &38.94 &35.93 &37.92 &41.71 &63.91 &47.37\\
  &SelfReg &26.08 &52.35 &53.89 &52.70 &35.11 &33.84 &37.46 &36.42 &50.99 &42.09 \\
  &MixStyle &35.98 &54.31 &57.35 &57.43 &37.14 &35.51 &39.31 &42.07 &57.84 &46.33 \\
  &MMD &38.67 &57.16 &61.49 &58.73 &\color{blue}\underline{40.65} &39.14 &38.69 &41.06 &\color{blue}\underline{71.48} &49.67 \\
  &SD &38.70 &51.06 &58.86 &\color{blue}\underline{62.21} &35.84 &40.05 &39.23 &\color{blue}\underline{44.34} &62.12 &48.04 \\
  &\textbf{DREAM} &\color{red}\textbf{43.84} &\color{red}\textbf{59.19} &\color{red}\textbf{66.09} &\color{red}\textbf{64.24} &39.59 &\color{red}\textbf{42.04} & \color{blue}\underline{40.49} &\color{red}\textbf{47.83} &68.12 &\color{red}\textbf{52.38} \\

  &\textbf{DREAM**} &\color{blue}\underline{39.68} &\color{blue}\underline{57.61} &\color{blue}\underline{64.48} &60.79&\color{red}\textbf{40.78} &\color{blue}\underline{40.10} & \color{red}\textbf{43.54} &43.80&\color{red}\textbf{72.42} &\color{blue}\underline{51.47} \\
  \midrule[1.3pt]
\end{tabular}
\end{threeparttable}
\end{table*}

\begin{figure*}
	\centering
	\begin{minipage}{0.325\linewidth}
		\centering
		\includegraphics[width=1\linewidth]{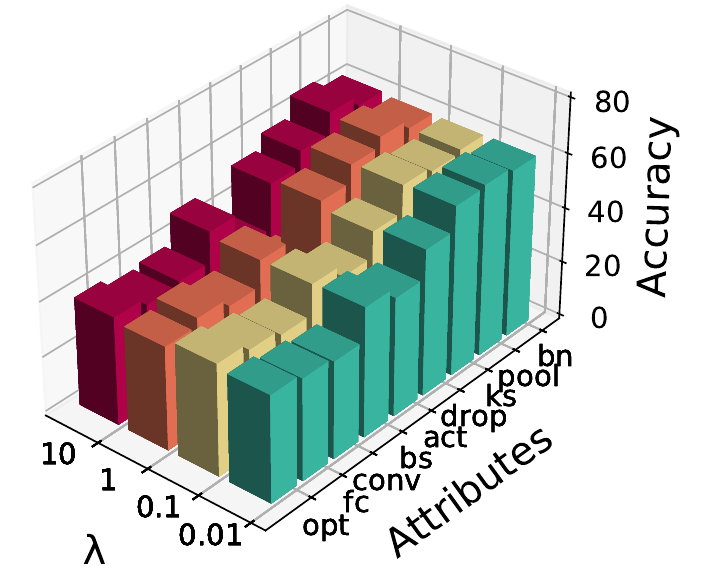}
		\label{fig:param_id2_P}
	\end{minipage}
	\begin{minipage}{0.325\linewidth}
		\centering
		\includegraphics[width=1\linewidth]{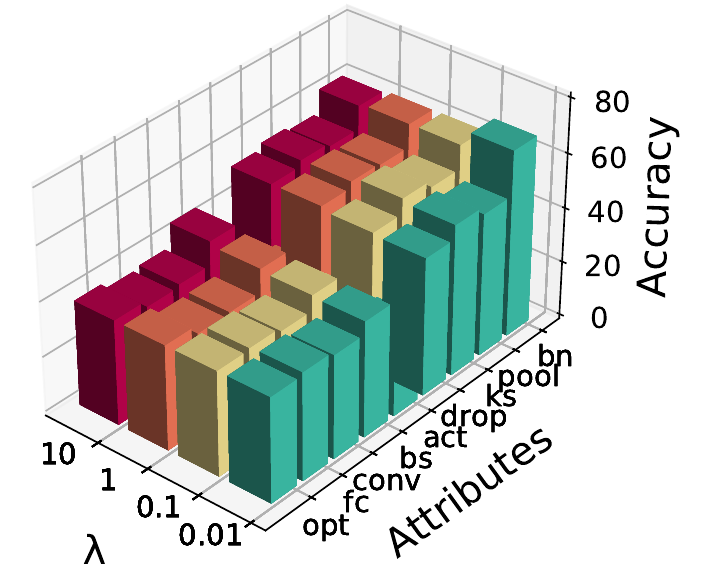}
		\label{fig:param_id2_C}
	\end{minipage}
	\begin{minipage}{0.325\linewidth}
		\centering
		\includegraphics[width=1\linewidth]{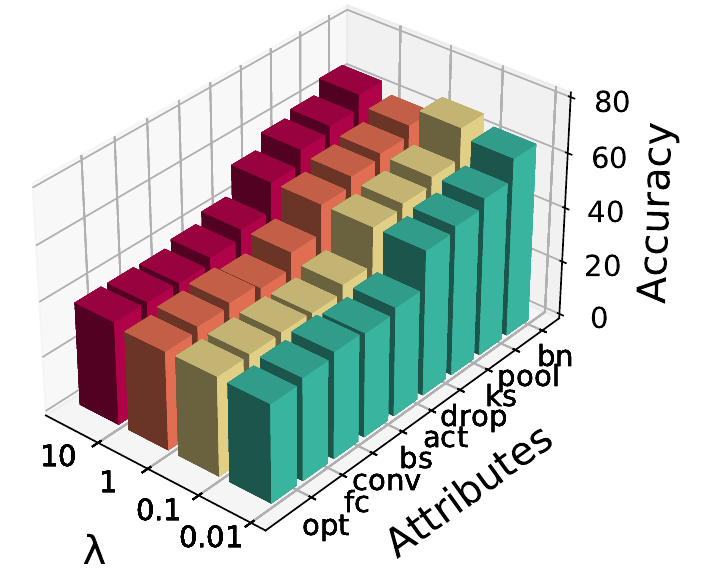}
		\label{fig:param_id2_S}
	\end{minipage}
	\\

	\caption{Sensitivity analysis of parameter $\lambda$  on PACS-modelset. From left to right, the results in the P split, C split and S split are shown, respectively.}
	\label{fig:sensitivity}
\end{figure*}
\begin{figure*}
	\centering
	\begin{minipage}{0.3\linewidth}
		\centering
		\includegraphics[width=1\linewidth]{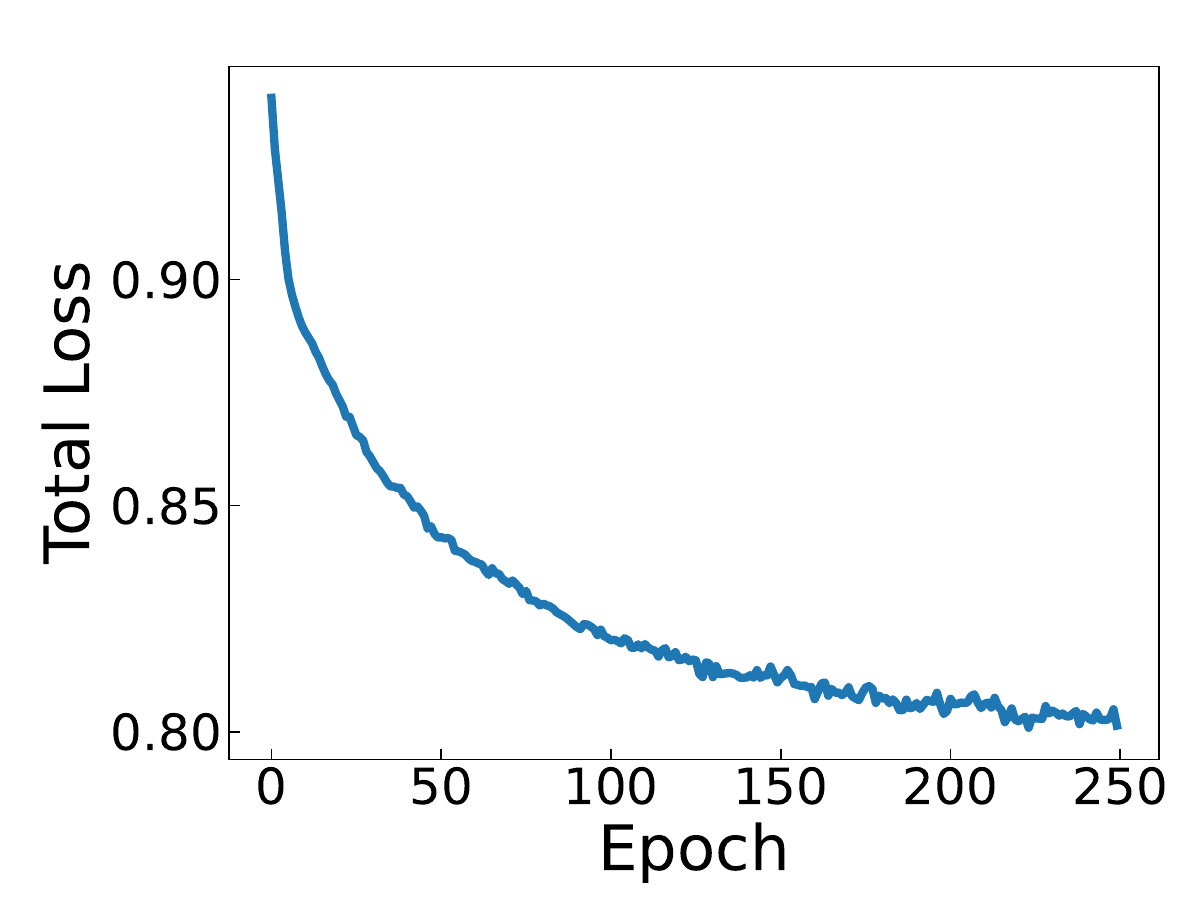}
		\label{fig:param_id2_P}
	\end{minipage}
	\begin{minipage}{0.3\linewidth}
		\centering
		\includegraphics[width=1\linewidth]{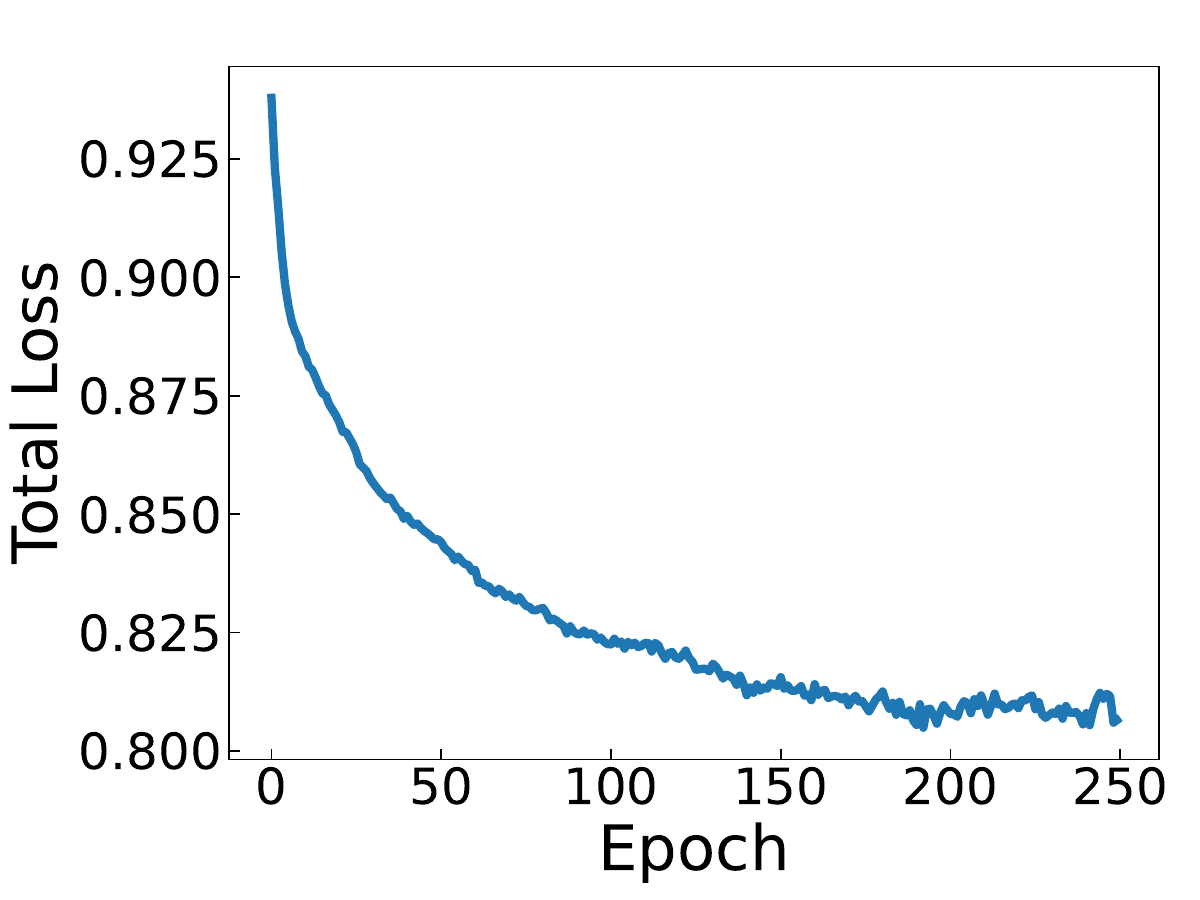}
		\label{fig:param_id2_C}
	\end{minipage}
	\begin{minipage}{0.3\linewidth}
		\centering
		\includegraphics[width=1\linewidth]{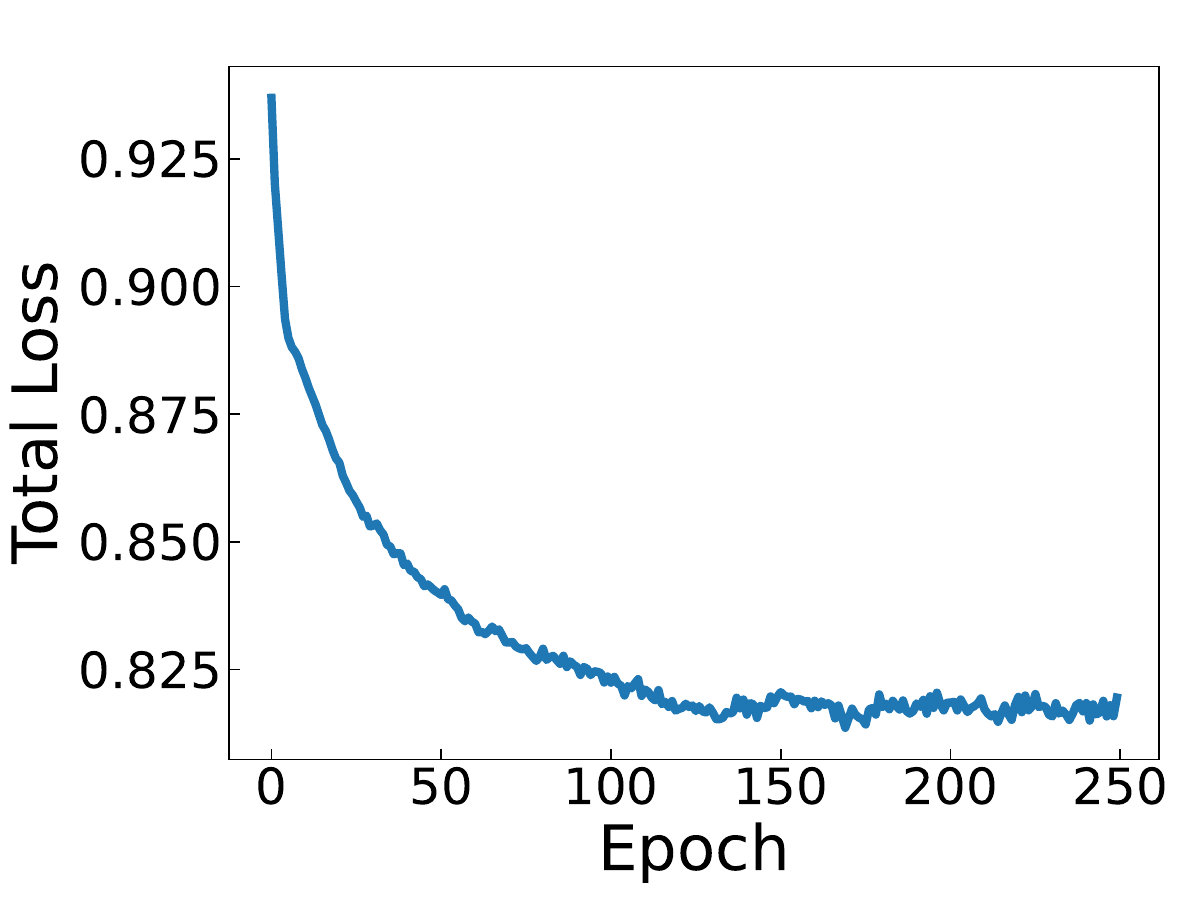}
		\label{fig:param_id2_S}
	\end{minipage}
	\caption{Convergence analysis. Loss curves of meta-classifier (Eq. \ref{equ:metace}) in the training phase on the three splits P, C, and S (from left to right) demonstrate the convergence of our method.}
	\label{fig:convergence}
\end{figure*}

\subsection{Visualization of Generated Feature Space}
To further verify the effectiveness of our method, we utilize t-SNE \cite{van2008visualizing} to visualize samples in the domain invariant feature space learned by the generator G of our framework. The visualization is carried out on PACS-modelset. We take C (cartoon) and P (photo) as source domains to train white-box models, and use S (sketch) as the unseen target domain to the train black-box model. As shown in Fig. \ref{fig:tsne_DREAM} a), samples from the three different domains are grouped into individual clusters at the 1st epoch. This illustrates their distributions are indeed different in the beginning. Distributions of source domains (C and P) become closer from epoch \#1 to \#5. Then, our method embeds features from the unseen domain (S) and the samples from the target domain also become closer to the source domains at the 10th epoch, indicating the generator is able to generalize an unseen domain into the feature space where the source domains are in. Finally, both source and target domains are transformed into an invariant feature space. For MMD and Mixstyle in Fig. \ref{fig:tsne_DREAM} b) and Fig. \ref{fig:tsne_DREAM} c), the distributions of features do not become closer as the training proceeds. For SelfReg in Fig. \ref{fig:tsne_DREAM} d), the features are pulled closer to some extent from epoch \#1 to epoch \#10, and part of samples from the target domain S indeed become closer to the source domains at the $100^{th}$ epoch. However, these feature distributions are not sufficiently tight. 

\begin{figure*}
	\centering
	\begin{minipage}{0.3\linewidth}
		\centering
		\includegraphics[width=1\linewidth]{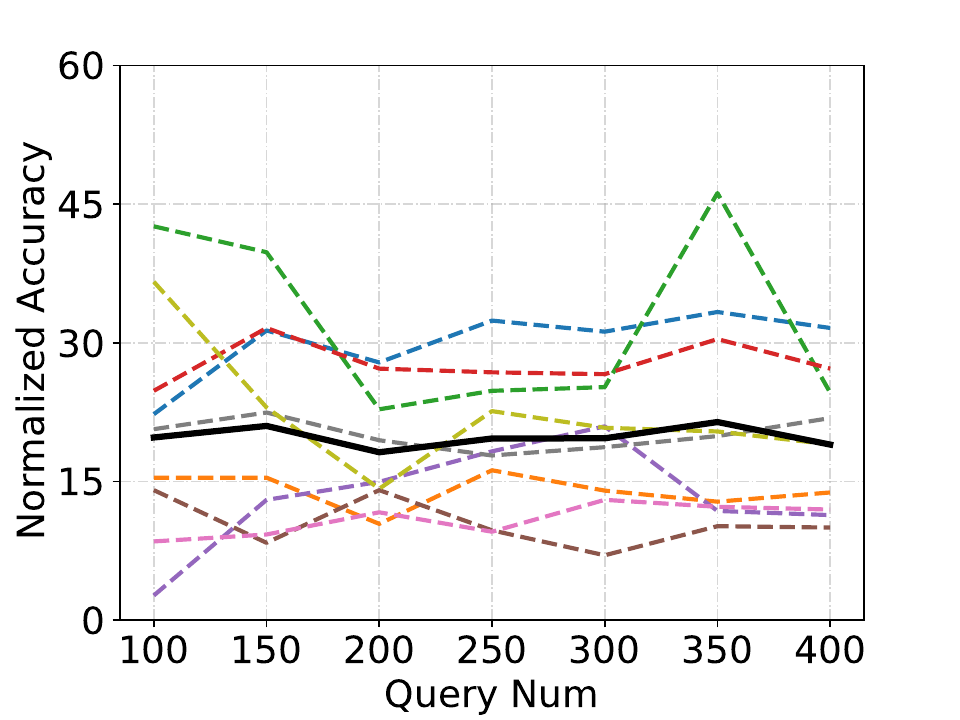}
		\label{fig:param_id2_P}
	\end{minipage}
	\begin{minipage}{0.3\linewidth}
		\centering
		\includegraphics[width=1\linewidth]{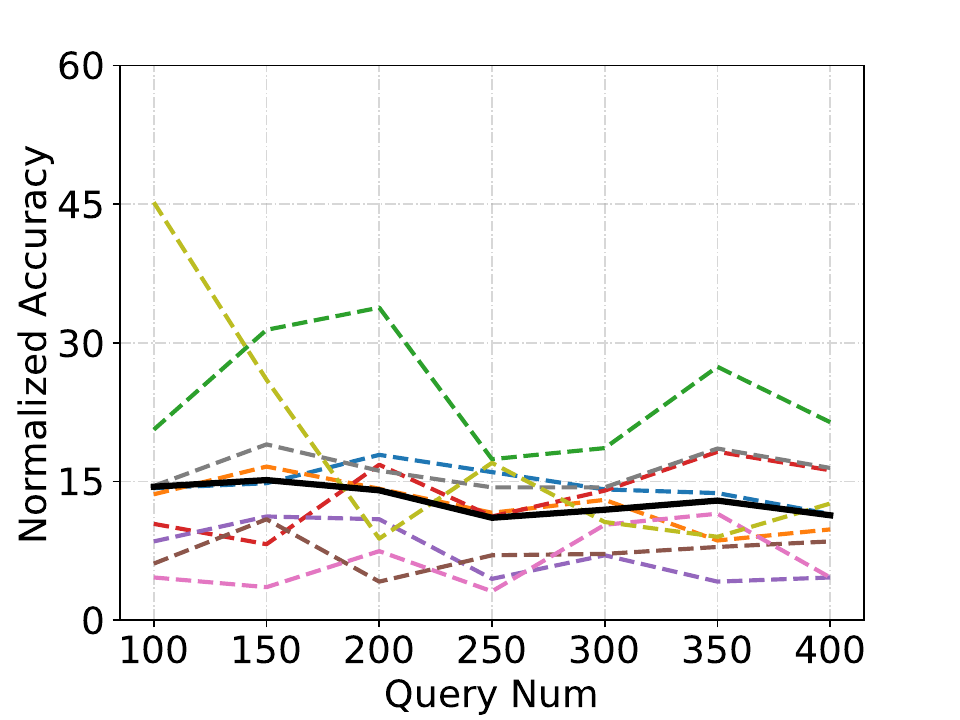}
		\label{fig:param_id2_C}
	\end{minipage}
	\begin{minipage}{0.3\linewidth}
		\centering
		\includegraphics[width=1\linewidth]{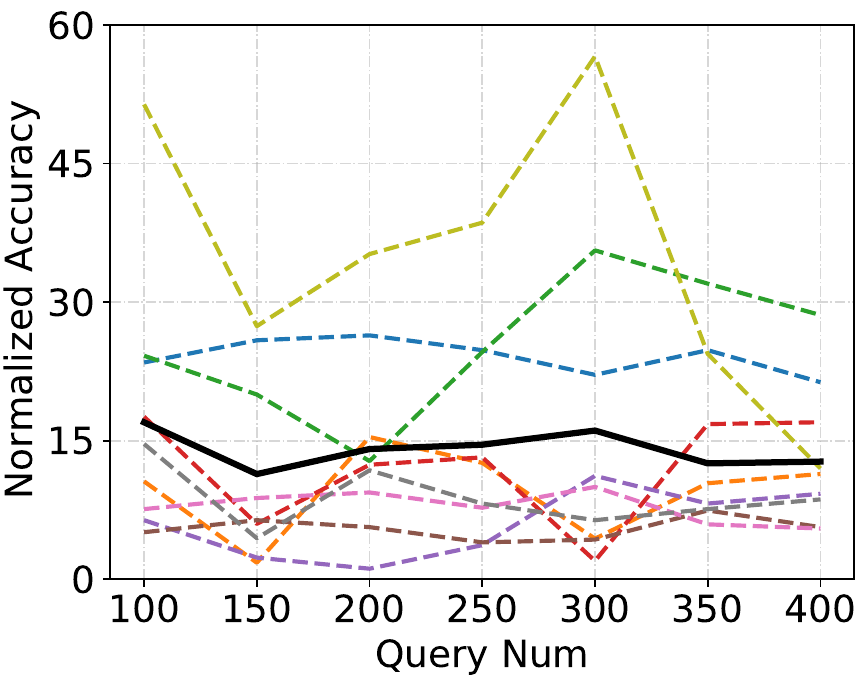}
		\label{fig:param_id2_S}
	\end{minipage}
	\begin{minipage}{0.08\linewidth}
		\centering
		\includegraphics[width=1\linewidth]{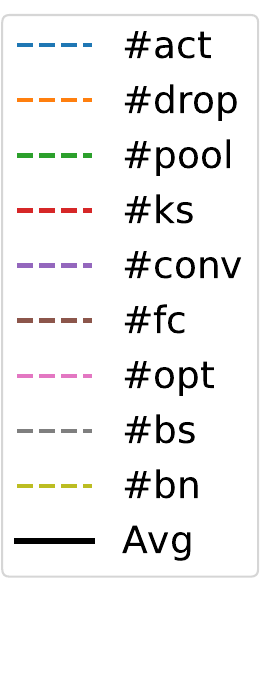}
		\label{fig:param_id2_S}
	\end{minipage}
	\caption{Performance against query number on PACS-modelset. From left to right, normalized accuracies in the P split, C split and S split are shown, respectively.}
	\label{fig:acc_query}
\end{figure*}

\begin{figure*}
	\centering
	\begin{minipage}{0.3\linewidth}
		\centering
		\includegraphics[width=1\linewidth]{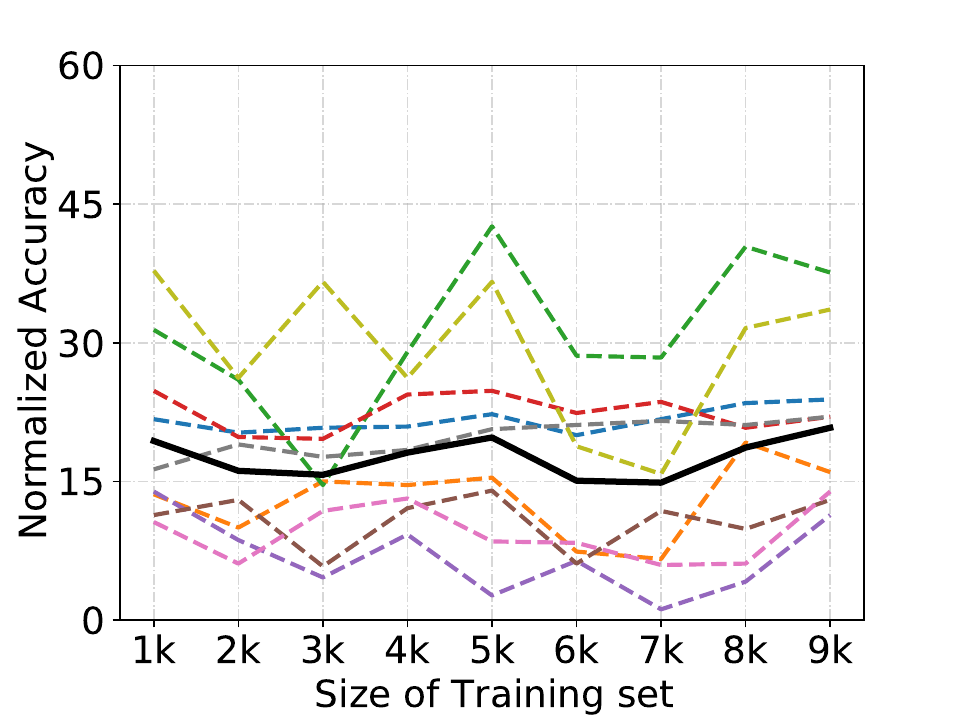}
		\label{fig:param_id2_P}
	\end{minipage}
	\begin{minipage}{0.3\linewidth}
		\centering
		\includegraphics[width=1\linewidth]{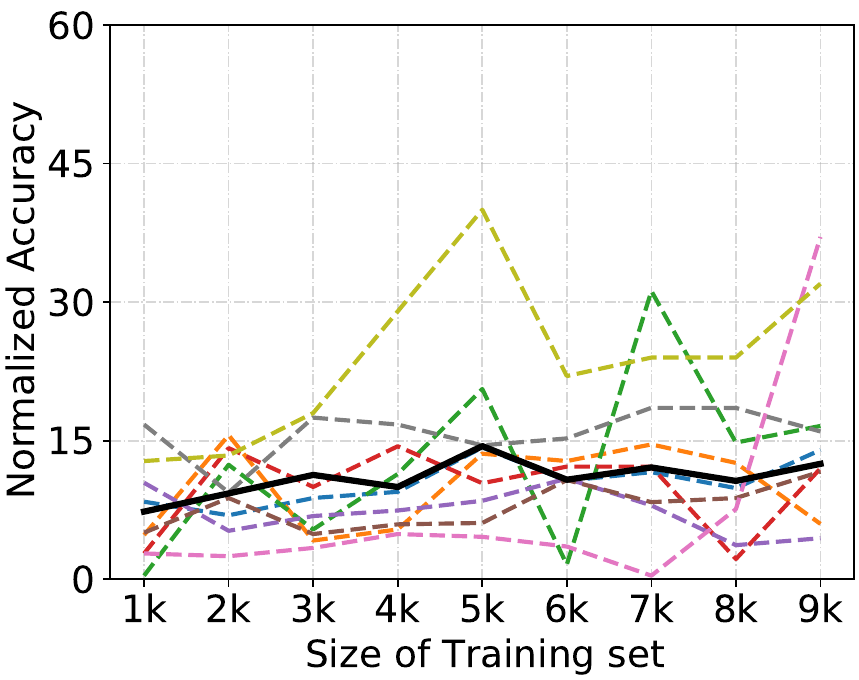}
		\label{fig:param_id3_C}
	\end{minipage}
	\begin{minipage}{0.3\linewidth}
		\centering
		\includegraphics[width=1\linewidth]{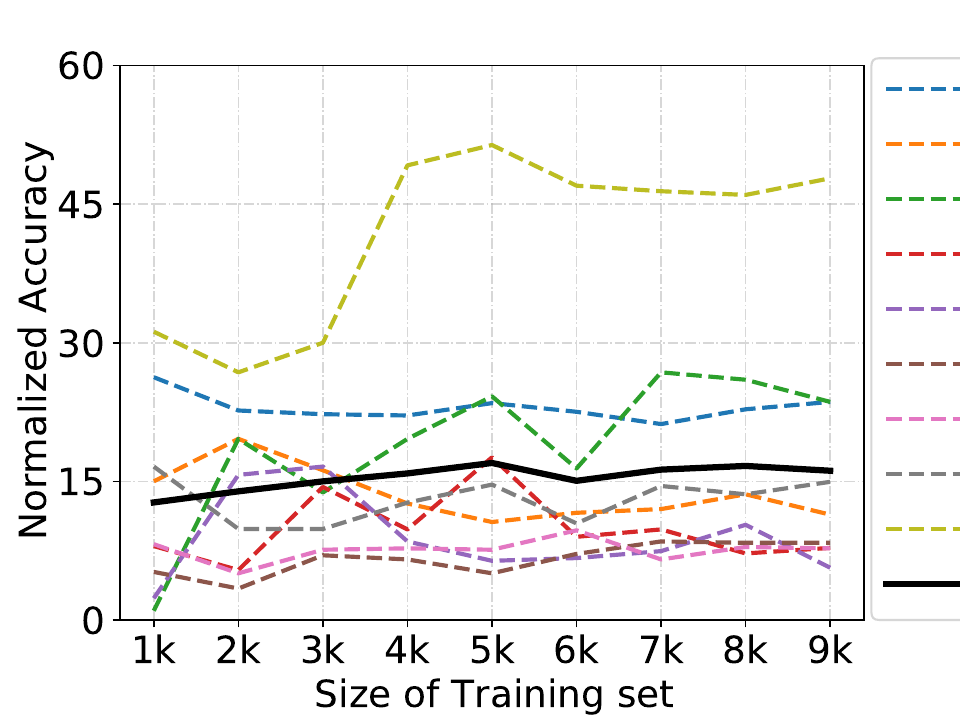}
		\label{fig:param_id1_S}
	\end{minipage}
	\begin{minipage}{0.08\linewidth}
		\centering
		\includegraphics[width=1\linewidth]{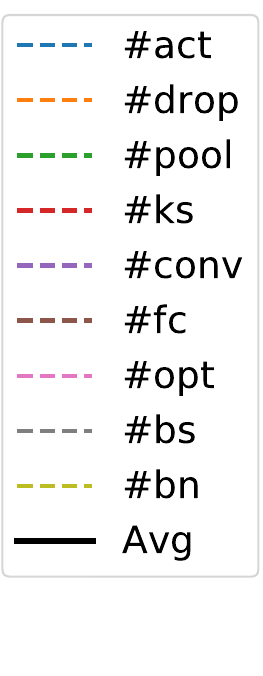}
		\label{fig:legend}
	\end{minipage}
	\caption{Performance against size of training set on PACS-modelset. From left to right, normalized accuracies in the P split, C split and S split are shown, respectively.}
	\label{fig:acc_size}
\end{figure*}

\begin{table*}[htbp]
\scriptsize
\caption{Distribution of attributes in domain P (Photo) of PACS-modelset and classification accuracy on the Photo validation set.}
\label{PACS-P}
\vspace{-2.5em}
\begin{centering}
\setlength{\tabcolsep}{0.3em}
\begin{tabular}{cc*{4}{c}c*{2}{c}c*{2}{c}c*{2}{c}c*{3}{c}c*{3}{c}c*{3}{c}c*{3}{c}c*{2}{c}}
\hspace{-0.8em} & \tabularnewline
 & \hspace{1em} & \multicolumn{4}{c}{act}  && \multicolumn{2}{c}{drop} & & \multicolumn{2}{c}{pool} && \multicolumn{2}{c}{ks} && \multicolumn{3}{c}{conv} && \multicolumn{3}{c}{fc} && \multicolumn{3}{c}{alg} && \multicolumn{3}{c}{bs} && \multicolumn{2}{c}{bn} \tabularnewline
 \cline{3-6}   \cline{8-9}    \cline{11-12}    \cline{14-15}  \cline{17-19} \cline{21-23} \cline{25-27} \cline{29-31} \cline{33-34}
\vspace{-1em} &  &  &  &  &  &  & & & \tabularnewline
 & \hspace{0.15em}& ReLU & ELU & PReLU & Tanh & \hspace{0.15em} & No & Yes & \hspace{0.15em} & No & Yes & \hspace{0.15em} & 3 & 5 & \hspace{0.15em} & 2 & 3 & 4 & \hspace{0.15em} & 2 & 3 & 4 & \hspace{0.15em} & SGD & ADAM & RMSprop &\hspace{0.15em} & 32 & 64 & 128 & \hspace{0.15em} & Yes & No \tabularnewline
\vspace{-1em} &   \tabularnewline
\cline{1-1} \cline{3-6}   \cline{8-9}    \cline{11-12}    \cline{14-15}  \cline{17-19} \cline{21-23} \cline{25-27} \cline{29-31} \cline{33-34}
\vspace{-1em} &   \tabularnewline
\cline{1-1} \cline{3-6}   \cline{8-9}    \cline{11-12}    \cline{14-15}  \cline{17-19} \cline{21-23} \cline{25-27} \cline{29-31} \cline{33-34}
\vspace{-0.8em} &   \tabularnewline
Ratio   && 24.7 & 25.8 & 24.5 & 25.0 && 50.9 & 49.1 && 50.3 & 49.7 && 50.6 & 49.4 && 33.5 & 33.2 & 33.3 && 33.2 & 32.8 & 34.0 && 32.4 & 34.0 & 33.7 && 33.6 & 33.9 & 32.6 && 49.8 & 50.2\tabularnewline
\cline{1-1} \cline{3-6}   \cline{8-9}    \cline{11-12}    \cline{14-15}  \cline{17-19} \cline{21-23} \cline{25-27} \cline{29-31} \cline{33-34}
\vspace{-0.8em} &   \tabularnewline
max    && 71.3 & 70.9 & 72.8 & 68.7 && 72.8 & 71.3 && 72.4 & 72.8 && 72.8 & 72.4 && 70.5 & 72.8 & 72.4 && 72.8& 71.1 & 72.4 && 72.4 & 70.9 & 72.8 && 71.3 & 72.8 & 72.4 && 68.3 & 72.8\tabularnewline
median &&  60.0 & 61.0 & 60.2 & 58.5 && 60.2 & 59.1 && 59.1 & 61.2 && 60.8 & 59.3 && 59.3 & 61.0 &  59.8 && 61.0 & 60.2 & 58.5 && 56.5 & 60.6 & 61.0 && 61.2 & 60.0 & 57.9 && 56.3& 63.6\tabularnewline
mean   && 58.0 & 59.7 &58.4 & 55.7 &&58.2 & 57.7 && 57.9 & 58.0 && 58.6 & 57.3 && 58.2 & 58.6 &  57.0 && 59.6 & 58.3 & 56.1 &&  54.6 & 60.2 & 58.9 && 59.7 & 58.1& 56.0 &&53.5 & 62.4\tabularnewline
min    && 26.0 & 26.2 & 26.2 & 25.6 && 25.6 & 26.0 && 25.6 & 25.8 && 26.2 & 25.6 && 26.0 & 25.8 &  25.6 && 26.0 & 25.8 & 25.6 && 25.8 & 26.2 & 25.6 && 26.2& 25.6 & 25.8 && 25.8 & 25.6\tabularnewline
\cline{1-1} \cline{3-6}   \cline{8-9}    \cline{11-12}    \cline{14-15}  \cline{17-19} \cline{21-23} \cline{25-27} \cline{29-31} \cline{33-34}
\end{tabular}

\par\end{centering}
\vspace{-1.5em}
\end{table*}

\begin{table*}[htbp]
\scriptsize
\caption{Distribution of attributes in domain C (Cartoon) of PACS-modelset and classification accuracy on the Cartoon validation set.}
\label{PACS-C}
\vspace{-2.5em}
\begin{centering}
\setlength{\tabcolsep}{0.3em}
\begin{tabular}{cc*{4}{c}c*{2}{c}c*{2}{c}c*{2}{c}c*{3}{c}c*{3}{c}c*{3}{c}c*{3}{c}c*{2}{c}}
\hspace{-0.8em} & \tabularnewline
 & \hspace{1em} & \multicolumn{4}{c}{act}  && \multicolumn{2}{c}{drop} & & \multicolumn{2}{c}{pool} && \multicolumn{2}{c}{ks} && \multicolumn{3}{c}{conv} && \multicolumn{3}{c}{fc} && \multicolumn{3}{c}{alg} && \multicolumn{3}{c}{bs} && \multicolumn{2}{c}{bn} \tabularnewline
 \cline{3-6}   \cline{8-9}    \cline{11-12}    \cline{14-15}  \cline{17-19} \cline{21-23} \cline{25-27} \cline{29-31} \cline{33-34}
\vspace{-1em} &  &  &  &  &  &  & & & \tabularnewline
 & \hspace{0.15em}& ReLU & ELU & PReLU & Tanh & \hspace{0.15em} & No & Yes & \hspace{0.15em} & No & Yes & \hspace{0.15em} & 3 & 5 & \hspace{0.15em} & 2 & 3 & 4 & \hspace{0.15em} & 2 & 3 & 4 & \hspace{0.15em} & SGD & ADAM & RMSprop &\hspace{0.15em} & 32 & 64 & 128 & \hspace{0.15em} & Yes & No \tabularnewline
\vspace{-1em} &   \tabularnewline
\cline{1-1} \cline{3-6}   \cline{8-9}    \cline{11-12}    \cline{14-15}  \cline{17-19} \cline{21-23} \cline{25-27} \cline{29-31} \cline{33-34}
\vspace{-1em} &   \tabularnewline
\cline{1-1} \cline{3-6}   \cline{8-9}    \cline{11-12}    \cline{14-15}  \cline{17-19} \cline{21-23} \cline{25-27} \cline{29-31} \cline{33-34}
\vspace{-0.8em} &   \tabularnewline
Ratio   && 24.9 & 26.6 & 24.9 & 23.6 && 51.0 & 49.0 && 50.7 & 49.3 && 50.6 & 49.4 && 34.5 & 33.4 & 32.1 && 34.3 & 32.9 & 32.3 && 30.5 & 36.1 & 33.4 && 34.4 & 34.0 & 31.6 && 46.4 & 53.5  \tabularnewline
\cline{1-1} \cline{3-6}   \cline{8-9}    \cline{11-12}    \cline{14-15}  \cline{17-19} \cline{21-23}\cline{25-27} \cline{29-31} \cline{33-34}
\vspace{-0.8em} &   \tabularnewline
max    && 73.0 & 71.0 & 71.7 & 69.9 && 71.7 & 73.0&& 71.6 & 73.0 && 73.0& 71.7 && 73.0 & 71.9 & 71.7 && 73.0 & 71.9 & 70.7 && 71.6 & 71.9& 73.0&& 73.0 & 71.7 & 71.7 && 70.9 & 73.0\tabularnewline
median && 62.5 & 63.3& 62.3 & 61.6 &&62.8 & 62.0 && 61.6 & 63.5&& 63.0& 61.7 && 60.9 & 63.2 &  63.2&& 63.0 & 62.8 & 61.4 && 61.0 & 62.8 & 62.9 && 63.0 & 62.3 & 61.6 && 59.9 & 64.5\tabularnewline
mean   && 60.5 & 61.7 & 61.1 & 59.4 &&61.2 & 60.2 && 60.6 & 60.8 && 61.5 & 60.1 && 60.1 & 61.2 &  61.0 && 61.8 & 60.9 &59.4 && 59.0 & 61.3 & 61.6 && 61.6 & 60.9 & 59.6 && 58.0 & 63.1\tabularnewline
min    && 25.7 & 25.1 & 26.2 & 25.8 && 25.1 & 25.8&& 26.2 & 25.1 && 25.8 & 25.1 && 25.1 & 25.7 &  27.0 && 25.7 & 25.8& 25.1 && 25.1& 27.2 & 28.8 && 25.8& 25.8 & 25.1 && 25.1 & 28.4\tabularnewline
\cline{1-1} \cline{3-6}   \cline{8-9}    \cline{11-12}    \cline{14-15}  \cline{17-19} \cline{21-23}\cline{25-27} \cline{29-31} \cline{33-34}
\end{tabular}
\par\end{centering}
\vspace{-1.5em}
\end{table*}

\begin{table*}[htbp]
\scriptsize
\caption{Distribution of attributes in domain S (Sketch) of PACS-modelset and classification accuracy on the Sketch validation set.}
\label{PACS-S}
\vspace{-2.5em}
\begin{centering}
\setlength{\tabcolsep}{0.3em}
\begin{tabular}{cc*{4}{c}c*{2}{c}c*{2}{c}c*{2}{c}c*{3}{c}c*{3}{c}c*{3}{c}c*{3}{c}c*{2}{c}}
\hspace{-0.8em} & \tabularnewline
 & \hspace{1em} & \multicolumn{4}{c}{act}  && \multicolumn{2}{c}{drop} & & \multicolumn{2}{c}{pool} && \multicolumn{2}{c}{ks} && \multicolumn{3}{c}{conv} && \multicolumn{3}{c}{fc} && \multicolumn{3}{c}{alg} && \multicolumn{3}{c}{bs} && \multicolumn{2}{c}{bn} \tabularnewline
 \cline{3-6}   \cline{8-9}    \cline{11-12}    \cline{14-15}  \cline{17-19} \cline{21-23} \cline{25-27} \cline{29-31} \cline{33-34}
\vspace{-1em} &  &  &  &  &  &  & & & \tabularnewline
 & \hspace{0.15em}& ReLU & ELU & PReLU & Tanh & \hspace{0.15em} & No & Yes & \hspace{0.15em} & No & Yes & \hspace{0.15em} & 3 & 5 & \hspace{0.15em} & 2 & 3 & 4 & \hspace{0.15em} & 2 & 3 & 4 & \hspace{0.15em} & SGD & ADAM & RMSprop &\hspace{0.15em} & 32 & 64 & 128 & \hspace{0.15em} & Yes & No \tabularnewline
\vspace{-1em} &   \tabularnewline
\cline{1-1} \cline{3-6}   \cline{8-9}    \cline{11-12}    \cline{14-15}  \cline{17-19} \cline{21-23} \cline{25-27} \cline{29-31} \cline{33-34}
\vspace{-1em} &   \tabularnewline
\cline{1-1} \cline{3-6}   \cline{8-9}    \cline{11-12}    \cline{14-15}  \cline{17-19} \cline{21-23} \cline{25-27} \cline{29-31} \cline{33-34}
\vspace{-0.8em} &   \tabularnewline
Ratio   && 25.7 & 27.3 & 25.8 & 21.2 && 51.5 & 48.5 && 49.3 & 50.7 && 50.7 & 49.3 && 34.7 & 33.5 & 31.8 && 34.9 & 32.9 & 32.3 && 32.0 & 35.8 & 32.2 && 34.3 & 33.6 & 32.1 && 46.0 & 54.0\tabularnewline
\cline{1-1} \cline{3-6}   \cline{8-9}    \cline{11-12}    \cline{14-15}  \cline{17-19} \cline{21-23} \cline{25-27} \cline{29-31} \cline{33-34}
\vspace{-0.8em} &   \tabularnewline
max    && 67.4 & 66.5 & 65.0 & 64.8 && 67.4 & 65.6 && 67.4 & 66.5&& 66.5 &  67.4 && 66.5 & 64.9 &  67.4&& 65.1 & 67.4 & 66.5 && 63.5 & 67.4 & 66.5 && 67.4 & 65.7 & 65.8 && 64.8 & 67.4\tabularnewline
median && 55.8 & 56.3 & 55.8& 54.0 && 56.9 & 54.2 && 54.7 & 56.9 &&  56.4 &  54.8 && 53.5 & 56.1 &  57.3 && 56.0 & 55.9& 55.1 && 54.0 & 56.9 & 55.6 &&57.1 & 54.9 & 54.7 && 55.1 & 56.3\tabularnewline
mean   && 53.9 & 54.4 & 53.8 & 52.0 && 54.9& 52.2 && 52.5 & 54.7 && 54.6 & 52.6 && 52.0 & 53.9 & 55.1 && 56.0 & 53.7 & 52.8 &&  51.5 & 55.6 & 53.4 && 55.2 & 52.8 & 52.7 && 53.6 & 51.5\tabularnewline
min    && 25.1 & 25.2 & 25.1 & 25.1 && 25.1 & 25.1 && 25.1 & 25.1 && 25.1 & 25.1 && 25.1 & 25.1&  25.1 && 25.1 & 25.5 & 25.1 && 25.1 &25.1 & 25.1 &&25.1& 25.1 & 25.1 && 25.1 & 25.1\tabularnewline
\cline{1-1} \cline{3-6}   \cline{8-9}    \cline{11-12}    \cline{14-15}  \cline{17-19} \cline{21-23} \cline{25-27} \cline{29-31} \cline{33-34}
\end{tabular}
\par\end{centering}
\vspace{-1.5em}
\end{table*}

\begin{table*}[htbp]
\scriptsize
\caption{Distribution of attributes in domain M (MNIST) of MEDU-modelset and classification accuracy on the MNIST validation set.}
\label{MEDU-M}
\vspace{-2.5em}
\begin{centering}
\setlength{\tabcolsep}{0.3em}
\begin{tabular}{cc*{4}{c}c*{2}{c}c*{2}{c}c*{2}{c}c*{3}{c}c*{3}{c}c*{3}{c}c*{3}{c}c*{2}{c}}
\hspace{-0.8em} & \tabularnewline
 & \hspace{1em} & \multicolumn{4}{c}{act}  && \multicolumn{2}{c}{drop} & & \multicolumn{2}{c}{pool} && \multicolumn{2}{c}{ks} && \multicolumn{3}{c}{conv} && \multicolumn{3}{c}{fc} && \multicolumn{3}{c}{alg} && \multicolumn{3}{c}{bs} && \multicolumn{2}{c}{bn} \tabularnewline
 \cline{3-6}   \cline{8-9}    \cline{11-12}    \cline{14-15}  \cline{17-19} \cline{21-23} \cline{25-27} \cline{29-31} \cline{33-34}
\vspace{-1em} &  &  &  &  &  &  & & & \tabularnewline
 & \hspace{0.15em}& ReLU & ELU & PReLU & Tanh & \hspace{0.15em} & No & Yes & \hspace{0.15em} & No & Yes & \hspace{0.15em} & 3 & 5 & \hspace{0.15em} & 2 & 3 & 4 & \hspace{0.15em} & 2 & 3 & 4 & \hspace{0.15em} & SGD & ADAM & RMSprop &\hspace{0.15em} & 32 & 64 & 128 & \hspace{0.15em} & Yes & No \tabularnewline
\vspace{-1em} &   \tabularnewline
\cline{1-1} \cline{3-6}   \cline{8-9}    \cline{11-12}    \cline{14-15}  \cline{17-19} \cline{21-23} \cline{25-27} \cline{29-31} \cline{33-34}
\vspace{-1em} &   \tabularnewline
\cline{1-1} \cline{3-6}   \cline{8-9}    \cline{11-12}    \cline{14-15}  \cline{17-19} \cline{21-23} \cline{25-27} \cline{29-31} \cline{33-34}
\vspace{-0.8em} &   \tabularnewline
Ratio   && 24.7 & 25.8 & 24.5 & 25.0 && 50.9 & 49.1 && 50.3 & 49.7 && 50.6 & 49.4 && 33.5 & 33.3 & 33.2 && 33.2 & 32.8 & 34.0 && 32.4 & 34.0 & 33.6 && 33.5 & 33.9 & 32.5 && 49.7 & 50.3\tabularnewline
\cline{1-1} \cline{3-6}   \cline{8-9}    \cline{11-12}    \cline{14-15}  \cline{17-19} \cline{21-23} \cline{25-27} \cline{29-31} \cline{33-34}
\vspace{-0.8em} &   \tabularnewline
max    && 99.3 & 99.2 & 99.3 & 99.2 && 99.3 & 99.3 && 99.3 & 99.2 && 99.2 & 99.3 && 99.1 & 99.2 &  99.3 && 99.3 & 99.2 & 99.3 && 99.2 & 99.3 & 99.2 && 99.3 & 99.3 & 99.2 && 99.1 & 99.3\tabularnewline
median && 98.6 & 98.6 & 98.6 & 98.4 && 98.5 & 98.6 && 98.5 & 98.6 && 98.5 & 98.6 && 98.4 & 98.6 &  98.6 && 98.3 & 98.2 & 98.3 && 98.4 & 98.6 & 98.5 && 98.6 & 98.5 & 98.5 && 98.4 & 98.7\tabularnewline
mean   && 98.4 & 98.5 & 98.4 & 98.3 &&98.3 & 98.5 && 98.3 & 98.4 && 98.3 & 98.5 && 98.4 & 98.6 &  98.6 && 98.6 & 98.5 & 98.5 &&  98.1 & 98.5 & 98.4 && 98.5 & 98.5 & 98.5 && 98.2 & 98.6\tabularnewline
min    && 91.6 & 94.3 & 36.9 & 73.5 && 36.9 & 63.8 && 63.8 & 36.9 && 63.8 & 36.9 && 63.8 & 91.6 &  36.9 && 92.8 & 63.8 & 36.9  && 36.9 & 92.8 & 90.0 && 90.0 & 92.8 & 36.9 && 36.9 & 63.8\tabularnewline
\cline{1-1} \cline{3-6}   \cline{8-9}    \cline{11-12}    \cline{14-15}  \cline{17-19} \cline{21-23} \cline{25-27} \cline{29-31} \cline{33-34}
\end{tabular}
\par\end{centering}
\vspace{-1.5em}
\end{table*}

\begin{table*}[htbp]
\scriptsize
\caption{Distribution of attributes in domain E (EMNIST) of MEDU-modelset and classification accuracy on the EMNIST validation set.}
\label{MEDU-E}
\vspace{-2.5em}
\begin{centering}
\setlength{\tabcolsep}{0.3em}
\begin{tabular}{cc*{4}{c}c*{2}{c}c*{2}{c}c*{2}{c}c*{3}{c}c*{3}{c}c*{3}{c}c*{3}{c}c*{2}{c}}
\hspace{-0.8em} & \tabularnewline
 & \hspace{1em} & \multicolumn{4}{c}{act}  && \multicolumn{2}{c}{drop} & & \multicolumn{2}{c}{pool} && \multicolumn{2}{c}{ks} && \multicolumn{3}{c}{conv} && \multicolumn{3}{c}{fc} && \multicolumn{3}{c}{alg} && \multicolumn{3}{c}{bs} && \multicolumn{2}{c}{bn} \tabularnewline
 \cline{3-6}   \cline{8-9}    \cline{11-12}    \cline{14-15}  \cline{17-19} \cline{21-23} \cline{25-27} \cline{29-31} \cline{33-34}
\vspace{-1em} &  &  &  &  &  &  & & & \tabularnewline
 & \hspace{0.15em}& ReLU & ELU & PReLU & Tanh & \hspace{0.15em} & No & Yes & \hspace{0.15em} & No & Yes & \hspace{0.15em} & 3 & 5 & \hspace{0.15em} & 2 & 3 & 4 & \hspace{0.15em} & 2 & 3 & 4 & \hspace{0.15em} & SGD & ADAM & RMSprop &\hspace{0.15em} & 32 & 64 & 128 & \hspace{0.15em} & Yes & No \tabularnewline
\vspace{-1em} &   \tabularnewline
\cline{1-1} \cline{3-6}   \cline{8-9}    \cline{11-12}    \cline{14-15}  \cline{17-19} \cline{21-23} \cline{25-27} \cline{29-31} \cline{33-34}
\vspace{-1em} &   \tabularnewline
\cline{1-1} \cline{3-6}   \cline{8-9}    \cline{11-12}    \cline{14-15}  \cline{17-19} \cline{21-23} \cline{25-27} \cline{29-31} \cline{33-34}
\vspace{-0.8em} &   \tabularnewline
Ratio   && 24.7 & 25.9 & 24.5 & 25.0 && 51.0 & 49.0 && 50.3 & 49.7 && 50.6 & 49.4 && 33.6 & 33.2 & 33.2 && 33.2 & 32.8 & 34.0 && 32.3 & 34.0 & 33.7 && 33.6 & 34.0 & 32.5 && 49.7 & 50.3 \tabularnewline
\cline{1-1} \cline{3-6}   \cline{8-9}    \cline{11-12}    \cline{14-15}  \cline{17-19} \cline{21-23} \cline{25-27} \cline{29-31} \cline{33-34}
\vspace{-0.8em} &   \tabularnewline
max    && 99.6 & 99.6 & 99.5 & 99.4 && 99.5 & 99.6 && 99.6 & 99.6 && 99.6 & 99.6 && 99.5 & 99.6 &  99.6 && 99.5 & 99.6 & 99.6 && 99.6 & 99.6 & 99.5 && 99.5 & 99.6 & 99.5 && 99.5 & 99.6\tabularnewline
median && 99.1 & 99.1 & 99.1 & 98.9 && 99.1 & 99.0 && 99.0 & 99.1 && 99.0 & 99.1 && 98.9 & 99.1 &  99.1 && 99.1 & 99.1 & 99.0 && 98.7 & 99.1 & 99.1 && 99.1 & 99.1 & 99.0 && 98.9 & 99.1\tabularnewline
mean   && 98.7 & 98.9 & 98.8 & 98.5 &&98.8 & 98.7 && 98.8 & 98.7 && 98.6 & 98.8 && 98.6 & 98.8 &  98.7 && 98.9 & 98.8 & 98.5 &&  98.2 & 99.0 & 99.0 && 99.0 & 98.8 & 98.4 && 98.4 & 99.0\tabularnewline
min    && 35.7 & 87.8 & 31.9 & 31.9 && 31.9 & 33.4 && 79.2 & 31.9 && 31.9 & 31.9 && 94.5 & 40.6 & 31.9 && 94.7 & 41.3 & 31.9 && 31.9 & 96.9 & 79.2 && 79.2 & 80.0 & 31.9 && 31.9 & 93.8\tabularnewline
\cline{1-1} \cline{3-6}   \cline{8-9}    \cline{11-12}    \cline{14-15}  \cline{17-19} \cline{21-23} \cline{25-27} \cline{29-31} \cline{33-34}
\end{tabular}
\par\end{centering}
\vspace{-1.5em}
\end{table*}

\begin{table*}[htbp]
\scriptsize
\caption{Distribution of attributes in domain D (DIDA) of MEDU-modelset and classification accuracy on the DIDA validation set.}
\label{MEDU-D}
\vspace{-2.5em}
\begin{centering}
\setlength{\tabcolsep}{0.3em}
\begin{tabular}{cc*{4}{c}c*{2}{c}c*{2}{c}c*{2}{c}c*{3}{c}c*{3}{c}c*{3}{c}c*{3}{c}c*{2}{c}}
\hspace{-0.8em} & \tabularnewline
 & \hspace{1em} & \multicolumn{4}{c}{act}  && \multicolumn{2}{c}{drop} & & \multicolumn{2}{c}{pool} && \multicolumn{2}{c}{ks} && \multicolumn{3}{c}{conv} && \multicolumn{3}{c}{fc} && \multicolumn{3}{c}{alg} && \multicolumn{3}{c}{bs} && \multicolumn{2}{c}{bn} \tabularnewline
 \cline{3-6}   \cline{8-9}    \cline{11-12}    \cline{14-15}  \cline{17-19} \cline{21-23} \cline{25-27} \cline{29-31} \cline{33-34}
\vspace{-1em} &  &  &  &  &  &  & & & \tabularnewline
 & \hspace{0.15em}& ReLU & ELU & PReLU & Tanh & \hspace{0.15em} & No & Yes & \hspace{0.15em} & No & Yes & \hspace{0.15em} & 3 & 5 & \hspace{0.15em} & 2 & 3 & 4 & \hspace{0.15em} & 2 & 3 & 4 & \hspace{0.15em} & SGD & ADAM & RMSprop &\hspace{0.15em} & 32 & 64 & 128 & \hspace{0.15em} & Yes & No \tabularnewline
\vspace{-1em} &   \tabularnewline
\cline{1-1} \cline{3-6}   \cline{8-9}    \cline{11-12}    \cline{14-15}  \cline{17-19} \cline{21-23} \cline{25-27} \cline{29-31} \cline{33-34}
\vspace{-1em} &   \tabularnewline
\cline{1-1} \cline{3-6}   \cline{8-9}    \cline{11-12}    \cline{14-15}  \cline{17-19} \cline{21-23} \cline{25-27} \cline{29-31} \cline{33-34}
\vspace{-0.8em} &   \tabularnewline
Ratio   && 24.7 & 26.3 & 24.8 & 24.1 && 51.0 & 49.0 && 50.2 & 49.9 && 50.9 & 49.1 && 34.2 & 33.3 & 32.5 && 33.9 & 32.8 & 33.3 && 31.7 & 35.4 & 32.9 && 33.9 & 33.9 & 32.1 && 47.6 & 52.4\tabularnewline
\cline{1-1} \cline{3-6}   \cline{8-9}    \cline{11-12}    \cline{14-15}  \cline{17-19} \cline{21-23} \cline{25-27} \cline{29-31} \cline{33-34}
\vspace{-0.8em} &   \tabularnewline
max    && 97.8 & 98.1 & 97.9 & 98.0 && 97.8 & 98.1 && 97.1 & 97.6 && 98.6 & 98.1 && 94.0 & 94.4 &  99.1 && 97.9 & 98.0 & 98.1 && 98.0 & 97.7 & 98.1 && 97.9 & 98.1 & 97.9 && 96.6 & 98.1\tabularnewline
median && 94.0 & 94.4 & 94.5 & 93.7 && 93.9 & 94.3 && 94.0 & 94.3 && 93.9 & 94.4 && 93.0 & 94.5 &  94.8 && 94.3 & 94.1 & 94.0 && 93.2 & 94.3 & 94.5 && 94.6 & 94.2 & 93.5 && 93.3 & 95.0\tabularnewline
mean   && 92.7 & 93.1 & 93.2 & 92.0 && 92.8 & 92.7 && 92.9 & 92.6 && 92.3 & 93.2 && 91.6 & 93.2 &  93.6 && 93.2 & 92.7 & 92.3 &&  90.4 & 93.9 & 93.8 && 93.8 & 93.0 & 91.3 && 91.1 & 94.2\tabularnewline
min    && 25.4 & 25.0 & 25.2 & 25.1 && 25.0 & 25.1 && 32.7 & 25.0 && 25.0 & 25.0 && 26.0 & 25.1 & 25.0 && 26.5 & 25.0 & 25.1 && 25.0 & 68.8 & 43.2 && 26.1 & 25.1 & 25.0 && 25.0 & 43.2\tabularnewline
\cline{1-1} \cline{3-6}   \cline{8-9}    \cline{11-12}    \cline{14-15}  \cline{17-19} \cline{21-23} \cline{25-27} \cline{29-31} \cline{33-34}
\end{tabular}
\par\end{centering}
\vspace{-1.5em}
\end{table*}

\begin{table*}[h]
\scriptsize
\caption{Distribution of attributes in domain U (USPS) of MEDU-modelset and classification accuracy on the USPS validation set.}
\label{MEDU-U}
\vspace{-2.5em}
\begin{centering}
\setlength{\tabcolsep}{0.3em}
\begin{tabular}{cc*{4}{c}c*{2}{c}c*{2}{c}c*{2}{c}c*{3}{c}c*{3}{c}c*{3}{c}c*{3}{c}c*{2}{c}}
\hspace{-0.8em} & \tabularnewline
 & \hspace{1em} & \multicolumn{4}{c}{act}  && \multicolumn{2}{c}{drop} & & \multicolumn{2}{c}{pool} && \multicolumn{2}{c}{ks} && \multicolumn{3}{c}{conv} && \multicolumn{3}{c}{fc} && \multicolumn{3}{c}{alg} && \multicolumn{3}{c}{bs} && \multicolumn{2}{c}{bn} \tabularnewline
 \cline{3-6}   \cline{8-9}    \cline{11-12}    \cline{14-15}  \cline{17-19} \cline{21-23} \cline{25-27} \cline{29-31} \cline{33-34}
\vspace{-1em} &  &  &  &  &  &  & & & \tabularnewline
 & \hspace{0.15em}& ReLU & ELU & PReLU & Tanh & \hspace{0.15em} & No & Yes & \hspace{0.15em} & No & Yes & \hspace{0.15em} & 3 & 5 & \hspace{0.15em} & 2 & 3 & 4 & \hspace{0.15em} & 2 & 3 & 4 & \hspace{0.15em} & SGD & ADAM & RMSprop &\hspace{0.15em} & 32 & 64 & 128 & \hspace{0.15em} & Yes & No \tabularnewline
\vspace{-1em} &   \tabularnewline
\cline{1-1} \cline{3-6}   \cline{8-9}    \cline{11-12}    \cline{14-15}  \cline{17-19} \cline{21-23} \cline{25-27} \cline{29-31} \cline{33-34}
\vspace{-1em} &   \tabularnewline
\cline{1-1} \cline{3-6}   \cline{8-9}    \cline{11-12}    \cline{14-15}  \cline{17-19} \cline{21-23} \cline{25-27} \cline{29-31} \cline{33-34}
\vspace{-0.8em} &   \tabularnewline
Ratio   && 25.1 & 26.2 & 25.0 & 23.7 && 51.0 & 49.0 && 49.1 & 50.9 && 50.9 & 49.1 && 33.5 & 33.4 & 33.1 && 33.3 & 32.7 & 34.0 && 32.9 & 34.8 & 32.3 && 33.5 & 34.0 & 32.5 && 48.7 & 51.3\tabularnewline
\cline{1-1} \cline{3-6}   \cline{8-9}    \cline{11-12}    \cline{14-15}  \cline{17-19} \cline{21-23} \cline{25-27} \cline{29-31} \cline{33-34}
\vspace{-0.8em} &   \tabularnewline
max    && 98.9 & 98.9 & 98.8 & 98.7 && 98.8 & 98.9 && 98.9 & 98.8 && 98.8 & 98.9 && 98.5 & 98.7 & 98.9 && 98.9 & 98.8 & 98.9 && 98.9 & 98.9 & 98.8 && 98.9 & 98.8 & 98.8 && 98.6 & 98.9\tabularnewline
median && 97.2 & 97.1 & 97.2 & 96.9 && 97.1 & 97.1 && 96.8 & 97.3 && 96.9 & 97.3 && 96.5 & 97.3 &  97.5 && 97.1 & 97.1 & 97.1 && 96.9 & 97.3 & 97.2 && 97.3 & 97.1 & 96.8 && 98.6 & 98.9\tabularnewline
mean   && 96.5 & 96.3 & 96.6 & 94.9 && 96.4 & 95.8 && 95.3 & 96.8 && 95.8 & 96.4 && 94.8 & 96.5 &  97.0 && 96.2 & 96.3 & 95.8 &&  96.1 & 96.5 & 95.7 && 96.8 & 96.3 & 95.2 && 96.7 & 95.5\tabularnewline
min    && 30.4 & 30.5 & 29.6 & 26.1 && 26.1 & 26.7 && 26.7 & 26.1 && 26.7 & 26.1 && 28.5 & 26.7 &  26.1 && 36.1 & 26.7 & 26.1 && 26.1 & 29.7 & 26.7 && 31.8 & 26.7 & 26.1 && 26.1 & 26.7\tabularnewline
\cline{1-1} \cline{3-6}   \cline{8-9}    \cline{11-12}    \cline{14-15}  \cline{17-19} \cline{21-23} \cline{25-27} \cline{29-31} \cline{33-34}
\end{tabular}
\par\end{centering}
\vspace{-1.5em}
\end{table*}

\subsection{Experiment on Domain Shift Scenario}
\textcolor{black}{We take two scenarios of domain shift into account. The first scenario involves a difference in the number of output labels between the training and testing processes. The second scenario involves variations in the attribute combinations of the models during the training and testing processes.}

1) We leave out the “dog” and “elephant” for each domain when training the white box models. Then, we train domain-agnostic classifiers using model outputs without the “dog” and “elephant” classes and then test domain-agnostic classifiers using model outputs that hold all classes. The trained model is denoted as DREAM*, where white-box models are trained with only five classes (except dog and elephant) and are tested on all classes. 
\textcolor{black}{As shown in Table \ref{table:ccs}, the average accuracy of DREAM* reaches 50.03\%. Although DREAM* experiences a slight decrease in accuracy compared to DREAM due to domain shift, it still outperforms other baselines.}

2) We study the case that the white-box model (for training) and the black-box model to be inferred have completely different attribute combinations. As we mentioned in Section \ref{subsec:dataset_construction}, there are $5,184$ combinations of model attributes in total. We randomly sample $3,000$, $1,000$, $1,000$ models as training, validation and testing sets, and none of the models has identical attribute combinations. \textcolor{black}{The resulting model is denoted as DREAM**. As shown in Table \ref{table:diff_arc}, DREAM** consistently outperforms other baselines on the above setting. The reason why DREAM** maintains its high performance is that each model attribute is treated as an individual classification problem. As a result, the domain shift caused by variation in attribute combinations does not significantly impact its overall performance.}

\subsection{Analysis}
\textbf{Sensitivity Analysis.}
We study the sensitivity of the trade-off parameter $\lambda$ in our final loss function  on the PACS-modelset. As shown in Fig. \ref{fig:sensitivity}, the results   for each model attribute do not show evident fluctuation when changing $\lambda$, suggesting that our proposed method is not sensitive to the choices of $\lambda$ in a wide range.

\textbf{Convergence Analysis}
We study the convergence of our algorithm on the PACS-modelset.
The curves of the meta-classifier's loss in the training phase are shown in Fig. \ref{fig:convergence}. For all three splits of domains (left to right), the loss decreases as the training proceeds and finally levels off.

\textbf{Query Number Analysis.}
Moreover, we study the performance of DREAM against the number of queries on PACS-modelset. Following \cite{oh2018towards2}, we use the normalized accuracy that is linearly scaled according to random choice. As shown in Fig. \ref{fig:acc_query}, with the increase of query numbers, the average performance does not improve but fluctuates, which means more queries do not necessarily provide more information for our DREAM framework.

\textbf{Analysis of Size of Training Dataset.}
We further study the performance of our method against the size of the training set on PACS-modelset. As shown in Fig. \ref{fig:acc_size}, we observe that the performance slightly fluctuates from the size of 1K to 5K, and does not consistently increase when the size increases. We suspect it can be attributed to the difficulty of our problem for domain-agnostic attribute inference of the black-box model, and the nature of the OOD problem, $i.e.$, the noise level increases as the size of the training set increases. It is worth studying further.
\section{Conclusion}
In this paper, we studied the problem of domain-agnostic reverse engineering towards the attributes of the black-box model with unknown domain data, and cast it as an OOD generalization problem.
We proposed a new framework, DREAM, which can predict the attributes of a black-box model with an arbitrary  domain, and explored to learn domain invariant features from probability outputs in the scenario of black-box attribute inference.
Extensive experimental results demonstrated the effectiveness of our method.


\bibliographystyle{IEEEtran}
\bibliography{TIFS_main}
\begin{IEEEbiography}[
{\includegraphics[width=1in,height=1.25in,clip,keepaspectratio]{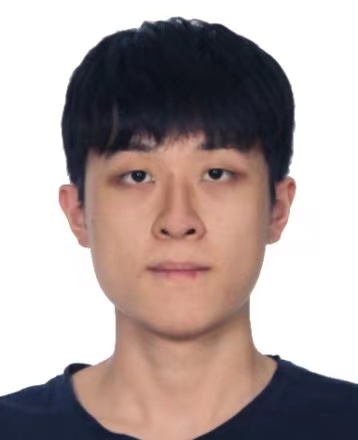}}
]{Rongqing Li}
Rongqing Li received the B.E. degree in Computer Science and Technology from Beijing University of Technology (BJUT) in 2021 and is currently pursuing the Ph.D. degree with the School of Computer Science and Technology, Beijing Institute of Technology (BIT), Beijing. His research interests include AI security, trajectory prediction, and computer vision.
\end{IEEEbiography}
\begin{IEEEbiography}[
{\includegraphics[width=1in,height=1.25in,clip,keepaspectratio]{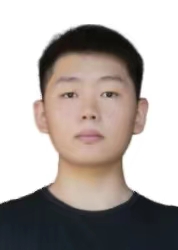}}
]{Jiaqi Yu}
received the B.E. degree in Computer Science and Technology from Nanjing University of Information Science and Technology (NUIST) in 2020. He is currently working for master degree in Computer Science and Technology at Beijing Institute of Technology (BIT). His research interests include recommendation system, computer vision and graph neural network.
\end{IEEEbiography}
\begin{IEEEbiography}[
{\includegraphics[width=1in,height=1.25in,clip,keepaspectratio]{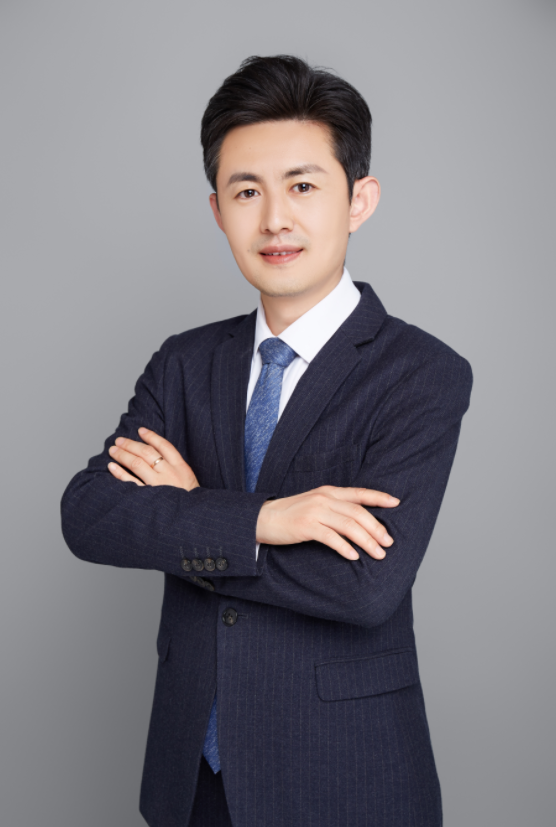}}
]{Changsheng Li}
received the B.E. degree from the University of Electronic Science and Technology of China (UESTC) in 2008 and the Ph.D. degree in pattern recognition and intelligent system from the Institute of Automation, Chinese Academy of Sciences, in 2013. During his Ph.D., he once studied as a Research Assistant with The Hong Kong Polytechnic University from 2009 to 2010. He is currently a Professor with the Beijing Institute of Technology. Before joining the Beijing Institute of Technology, he worked with IBM Research, China, Alibaba Group, and UESTC. He has more than 60 refereed publications in international journals and conferences, including IEEE TRANSACTIONS ON PATTERN ANALYSIS AND MACHINE INTELLIGENCE, INTERNATIONAL JOURNAL OF COMPUTER VISION, IEEE TRANSACTIONS ON IMAGE PROCESSING, IEEE TRANSACTIONS ON NEURAL NETWORKS AND LEARNING SYSTEMS, IEEE TRANSACTIONS ON COMPUTERS, IEEE TRANSACTIONS ON MULTIMEDIA, PR, CVPR, KDD, AAAI, IJCAI, CIKM, MM, and ICMR. His research interests include machine learning, data mining, and computer vision. He won the National Science Fund for Excellent Young Scholars in 2021.
\end{IEEEbiography}
\begin{IEEEbiography}[
{\includegraphics[width=1in,height=1.25in,clip,keepaspectratio]{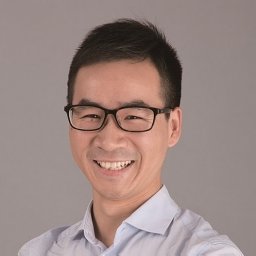}}
]{Wenhan Luo}
is currently an Associate Professor in Sun Yat-sen University. Prior to that, he worked as a research scientist for Tencent and Amazon. He has published over 60 papers in top conferences and leading journals, including ICML, CVPR, ICCV, ECCV, ACL, AAAI, ICLR, TPAMI, IJCV, TIP, etc. He also has been reviewer, senior PC member and Guest Editor for several prestigious journals and conferences. His research interests include several topics in computer vision and machine learning, such as image/video synthesis, image/video quality restoration, reinforcement learning. He received the Ph.D. degree from Imperial College London, UK, 2016, M.E. degree from Institute of Automation, Chinese Academy of Sciences, China, 2012 and B.E. degree from Huazhong University of Science and Technology, China, 2009.
\end{IEEEbiography}
\begin{IEEEbiography}[
{\includegraphics[width=1in,height=1.25in,clip,keepaspectratio]{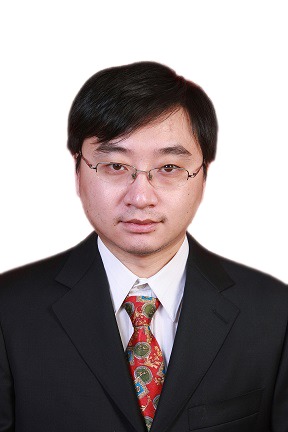}}
]{Ye Yuan} 
received the B.S., M.S., and Ph.D. degrees in computer science from Northeastern University in 2004, 2007, and 2011, respectively. He is currently a Professor with the Department of Computer Science, Beijing Institute of Technology, China. He has more than 100 refereed publications in international journals and conferences, including VLDBJ, IEEE TRANSACTIONS ON PARALLEL AND DISTRIBUTED SYSTEMS, IEEE TRANSACTIONS ON KNOWLEDGE AND DATA ENGINEERING,SIGMOD, PVLDB, ICDE, IJCAI, WWW, and KDD. His research interests include graph embedding, graph neural networks, and social network analysis. He won the National Science Fund for Excellent Young Scholars in 2016.
\end{IEEEbiography}
\begin{IEEEbiography}[
{\includegraphics[width=1in,height=1.25in,clip,keepaspectratio]{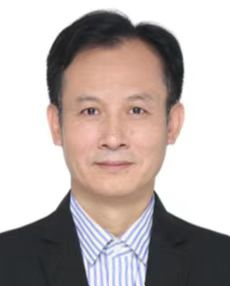}}
]{Guoren Wang}
received the B.S., M.S., and Ph.D. degrees in computer science from Northeastern University, Shenyang, in 1988, 1991, and 1996, respectively. He is currently a Professor with the School of Computer Science and Technology, Beijing Institute of Technology, Beijing, where he has been the Dean since 2020. He has more than 300 refereed publications in international journals and conferences, including VLDBJ, IEEE TRANS-ACTIONS ON PARALLEL AND DISTRIBUTED SYSTEMS, IEEE TRANSACTIONS ON KNOWLEDGE AND DATA ENGINEERING, SIGMOD, PVLDB, ICDE, SIGIR, IJCAI, WWW, and KDD. His research interests include data mining, database, machine learning, especially on highdimensional indexing, parallel database, and machine learning systems. He won the National Science Fund for Distinguished Young Scholars in 2010 and was appointed as the Changjiang Distinguished Professor in 2011.
\end{IEEEbiography}
\end{document}